\documentclass[10pt,journal,compsoc]{IEEEtran}
\ifCLASSOPTIONcompsoc
  \usepackage[nocompress]{cite}
\else
  \usepackage{cite}
\fi
\ifCLASSINFOpdf
\else
\fi
\usepackage{amssymb}
\usepackage{amsmath}
\usepackage{subfigure}
\usepackage{graphicx}
\usepackage{url}
\usepackage{makecell}
\usepackage{multirow}
\usepackage{color}

\newtheorem{theorem}{Theorem}[section]

\begin{document}

%
\title{Joint Active Learning with Feature Selection via CUR Matrix Decomposition}
%
%
%
%

\author{Changsheng~Li,
        Xiangfeng~Wang,
        Weishan~Dong,
        Junchi~Yan,~\IEEEmembership{Member,~IEEE,}
        Qingshan~Liu,~\IEEEmembership{Senior Member,~IEEE,}
        and~Hongyuan~Zha
\IEEEcompsocitemizethanks{

\IEEEcompsocthanksitem C. Li is with School of Computer Science and Engineering and Big Data Research Center, University of Electronic Science and Technology of China. He is also with AI Research Lab of Youedata. E-mail: lichangsheng@uestc.edu.cn
\IEEEcompsocthanksitem X. Wang is with Shanghai Key Lab for Trustworthy Computing, School of Computer Science and Software Engineering, East China Normal University, China. E-mail: xfwang@sei.ecnu.edu.cn
\IEEEcompsocthanksitem W. Dong is with Baidu Search. E-mail: dongweishan@baidu.com
\IEEEcompsocthanksitem J. Yan is with School of Electronic, Information and Electrical Engineering, Shanghai Jiao Tong University, China. E-mail: yanesta13@163.com
\IEEEcompsocthanksitem Q. Liu is with  the B-DAT Laboratory at the school of Information and Control, Nanjing University of Information Science and Technology, China. E-mail: qsliu@nuist.edu.cn
\IEEEcompsocthanksitem H. Zha is with School of Computational Science and Engineering, Georgia Institute of Technology, Atlanta, USA.
E-mail: zha@cc.gatech.edu
}
}

%
%

\markboth{Submitted to IEEE TRANSACTIONS ON PATTERN ANALYSIS AND MACHINE INTELLIGENCE}%
{Changsheng \MakeLowercase{\textit{et al.}}: Joint Active Learning with Feature Selection via CUR Matrix Decomposition}
%



\IEEEtitleabstractindextext{%
\begin{abstract}
This paper presents an unsupervised learning approach for simultaneous sample and feature selection, which is in contrast to existing works which mainly tackle these two problems separately. In fact the two tasks are often interleaved with each other: noisy and high-dimensional features will bring adverse effect on sample selection, while informative or representative samples will be beneficial to feature selection. Specifically, we propose a framework to jointly conduct active learning and feature selection based on the CUR matrix decomposition. From the data reconstruction perspective, both the selected samples and features can best approximate the original dataset respectively, such that the selected samples characterized by the features are highly representative. In particular, our method runs in one-shot without the procedure of iterative sample selection for progressive labeling. Thus, our model is especially suitable when there are few labeled samples or even in the absence of supervision, which is a particular challenge for existing methods. As the joint learning problem is NP-hard, the proposed formulation involves a convex but non-smooth optimization problem. We solve it efficiently by an iterative algorithm, and prove its global convergence. Experimental results on publicly available datasets corroborate the efficacy of our method compared with the state-of-the-art.
\end{abstract}

}

\maketitle

\IEEEdisplaynontitleabstractindextext

%
\IEEEpeerreviewmaketitle
\IEEEraisesectionheading{\section{Introduction}\label{sec:introduction}}
In many real-life machine learning tasks, unlabeled data is often readily available whereas labeled data are scarce.
Building powerful predictive models generally requires domain experts to manually annotate samples -- an expensive and time-consuming procedure. Active learning \cite{freund1997selective} provides a means to alleviate this problem by carefully selecting samples to be labeled by experts. Typically, the active learning algorithms prefer to query those unlabeled samples that can most improve prediction performance if samples were labeled and used as training data. In this way, the active learner aims to pick as few samples as possible to label for minimizing annotating cost, while accurate supervised learning models can be built by the labeled data.

In the past decade, lots of active learning algorithms have been proposed  \cite{cohn1996active,li2012multi,elhamifar2013convex,zhang2015bidirectional}, and have been successfully applied to a variety of problems in computer vision \cite{qi2008two,jain2009active,joshi2009multi,vijayanarasimhan2010far,liang2014beyond,vijayanarasimhan2014large}.
Broadly speaking, existing methods for actively selecting unlabeled samples for labeling can be categorized into two main groups \cite{wang2013querying}. The first group aims to select samples being of informativeness, where informativeness measures the ability of a sample in reducing the uncertainty of a statistical model \cite{huang2010active}. Common approaches include uncertainty sampling \cite{lewis1994sequential,yang2014multi}, query by committee \cite{freund1997selective}, and empirical risk minimization \cite{roy2001toward}. These algorithms are implemented iteratively, where a model is learned with the existing labeled data and new samples are chosen to be labeled based on the learned model. Since training often needs a large number of labeled data to avoid sampling bias, the methods above should be used after sufficient labeled samples are collected \cite{nie2013early}. The second group aims at querying samples being of representativeness, where representativeness measures if a sample well represents the overall input patterns of unlabeled data \cite{huang2010active,yu2006active,Chattopadhyay2012,cai2012manifold,nie2013early,hu2013active}. Contrasted with the first group, these methods are one-shot and non-iterative to select samples. Such methods are usually applied when there is no initial labeled data.

Although active learning has been well studied for years, it still has some issues in many real-world scenarios. For example, the sample is often characterized by high-dimensional features, and some of features are often noisy or irrelevant. These noisy or irrelevant features bring adverse influence on selecting informative or representative samples. Moreover, after querying samples, some supervised learning models, such as decision tree, are often trained based on these labeled data for various applications. However, high-dimensional features significantly increase the time and space requirements for model training. Meanwhile, when only limited labeled samples are available, it is difficult to guarantee reliable model parameter estimates in a high-dimensional feature space. One may state that, if we apply some state-of-the-art feature selection techniques \cite{Li2016Feature}, such as SPEC \cite{zhao2007spectral}, to learn a low-dimensional representation before active learning, these problems might be solved. To a certain extent, this is helpful for active learning to some extent; while common feature selection techniques and active learning algorithms are independent in designing, directly combining them usually cannot guarantee obtaining the optimal results.
Taking a feature selection method, Laplacian Score \cite{he2005laplacian}, and an active learning algorithm, RRSS \cite{nie2013early}, as an example, Laplacian Score aims to select a feature subset for preserving the local structure of data, while RRSS aims at finding an optimal sample subset to reconstruct data with the minimal error in the Euclidean space. Thus, it is hard to say that after performing Laplacian Score, the selected feature subset is the most suitable one for RRSS.
In light of this, it will benefit from devising a principled model for incorporating active learning and feature selection in a unified fashion. Recently, Joshi and Xu \cite{joshiactive} presented an active learning method with integrated feature selection based on linear kernel SVMs and GainRatio. Raghavan et al. \cite{raghavan2006active} intended to use human feedback on both features and samples for active learning. Kong et al. \cite{kong2011dual} proposed a dual feature and sample selection method in the context of graph classification.
Bilgic \cite{bilgic2012combining} proposed a dynamic dimensionality reduction algorithm that determined the appropriate number of dimensions for each active learning iteration. As all of the above three algorithms are implemented iteratively, and need to train models for querying in each iteration, they are suitable to work in the scenarios of the first group of active learning methods. {Differing from these iterative methods, we focus on studying the problem of the second active learning group, the case when no initial labeled samples are available, by jointly learning important features and samples. This is an unsupervised learning problem, which is harder due to the absence of labels that would guide the search for relevant information.}

In this paper, we present a unified view of \underline{A}ctive \underline{L}earning and \underline{F}eature \underline{S}election, called ALFS, inspired by the approximation method for CUR matrix decomposition. The main contributions of this paper are:
\begin{itemize}
\item[\romannumeral1)]
To our knowledge, this is the first work presenting a unified view for one-shot active learning and feature selection. This is important for real-world applications as it dispenses with any label effort unlike progressive interactive labeling active learning methods.
\item[\romannumeral2)]
This work is the first to formulate and build the natural connection between CUR decomposition and simultaneous sample and feature selection.
\item[\romannumeral3)]
 We devise a novel model and convex optimization algorithm to solve the one-shot sample and feature learning problem.
\item[\romannumeral4)]
 The convergence of the proposed iterative algorithm is theoretically proved, and extensive empirical results demonstrate the advantages of our approach.
\end{itemize}

%

{\bf Notations}. In this paper, matrices are written as boldface uppercase letters and vectors are written as boldface lowercase letters.
Given a matrix $\mathbf{P}$, we denote its $(i,j)$-th entry, $i$-th row, $j$-th column as $\mathbf{P}_{ij}$, $\mathbf{p}^i$, $\mathbf{p}_j$, respectively. The only vector norm used is the $l_2$ norm, denoted by $\|\!\cdot\!\|_2$. A variety of norms on matrices will be used. The $l_1$, $l_{2,1}$, $l_\infty$ norms of a matrix are defined by $\|\mathbf{P}\|_1=\sum_{i,j}|\mathbf{P}_{ij}|$, $\|\mathbf{P}\|_{2,1}=\sum_{i=1}^m\sqrt{\sum_{j=1}^n\mathbf{P}_{ij}^2}=\sum_{i=1}^m\|\mathbf{p}^i\|_2$, and $\|\mathbf{P}\|_\infty=\max_{i,j}|\mathbf{P}_{ij}|$, respectively.
The quasi-norm $l_{2,0}$ norm of a matrix $\mathbf{P}$ is defined as the number of the non-zero rows of $\mathbf{P}$, denoted by $\|\mathbf{P}\|_{2,0}$.
The Frobenius norm is denoted by $\|\mathbf{P}\|_F$. The Euclidean inner product between two matrices is $\langle \mathbf{P},\mathbf{Q}\rangle= tr(\mathbf{P}^T\mathbf{Q})$,
where $\mathbf{P}^T$ is the transpose of the matrix $\mathbf{P}$ and $tr(\cdot)$ is the trace of a matrix. The rank of a matrix is denoted by ${rank}(\cdot)$.
\section{RELATED WORK}
As described, the work most related to our proposed approach is the second group of active learning methods (discussed in section 1) that aim to select the most representative samples in the absence of supervision. In this section, we will briefly provide a review of the approaches of this group. Among them, the most popular one is the Transductive Experimental Design (TED) \cite{yu2006active}.  TED aims to find a representative sample subset from the unlabeled dataset, such that the dataset can be best approximated by linear combinations of the selected samples.
Since this optimization problem is NP-hard, \cite{yu2006active} proposed a suboptimal sequential optimization algorithm and a non-greedy optimization algorithm to solve it, respectively.

Following TED, more active learning algorithms have been developed.
Cai and He \cite{cai2012manifold} extended TED to choose samples by utilizing a nearest neighbor graph to capture the intrinsic local manifold structure, where the graph Laplacian is incorporated into a manifold adaptive kernel space.
Zhang et al. \cite{zhang2011active} adopted the idea from Locally Linear Embedding (LLE) \cite{roweis2000nonlinear} to find the reconstruction coefficients. They represented each sample by a linear combination of its neighbors, which can faithfully preserve the local geometrical structure of the data.
Similar to \cite{zhang2011active}, Hu et al. \cite{hu2013active} incorporated the local geometrical information into the active learning process. Specifically, they introduced a regularization term to make nearer neighbors have outsized effect on the linear reconstruction of a data point, and severely penalized selected samples distant from the reconstructed sample.
Nie et al. \cite{nie2013early} proposed a novel method to relax the objective of TED to an efficient convex formulation, and utilized the robust sparse representation loss function to reduce the effect of outliers.
\section{PROPOSED METHOD}
Given an unlabeled dataset $\mathbf{X}=[\mathbf{x}_1,\ldots,\mathbf{x}_n]\in \mathbb{R}^{d \times n}$, our goal is to pick out $m\ (m < n)$ samples for user labeling, and meanwhile simultaneously select $r\ (r < d)$ features as the new feature representation,  such that the potential performance is maximized when the model is trained based on the selected $m$ labeled samples under the new representation.
This is a more challenging problem than traditional representativeness based active learning problems, because selecting $m$ samples to best approximate $\mathbf{X}$ often leads to an NP-hard problem \cite{yu2006active}, and finding $r$ features as the most representative feature subset is also NP-hard \cite{he2011variance}.

\subsection{Active Learning and Feature Selection via Matrix Decomposition}
Inspired by the CUR matrix decomposition \cite{boutsidis2014optimal,mahoney2009cur,drineas2008relative,wang2013improving}, we propose a unified framework to find the most representative samples and features.
To make this paper self-contained, we first introduce CUR matrix factorization.
\newtheorem{definition}{Definition}[section]
\begin{definition}\label{def1}
Given $\mathbf{X}\in \mathbb{R}^{d\times n}$ of rank $\rho$ = $rank(\mathbf{X})$, rank parameter $k<\rho$, and accuracy parameter $0 <\varepsilon < 1$, the CUR factorization for $\mathbf{X}$ aims to find $\mathbf{C}\in \mathbb{R}^{d\times m}$ with $m$ columns from $\mathbf{X}$, $\mathbf{R}\in \mathbb{R}^{r\times n}$ with $r$ rows of $\mathbf{X}$, and $\mathbf{U}\in \mathbb{R}^{m\times r}$, with $m$, $r$, and $rank(\mathbf{U})$ being as small as possible, such that $\mathbf{X}$ is reconstructed within relative-error:
\begin{align}\label{svd1}
\|\mathbf{X}-\mathbf{C}\mathbf{U}\mathbf{R}\|_F^2\leq (1+\varepsilon)\|\mathbf{X}-\mathbf{X}_k\|_F^2,
\end{align}
\end{definition}
where $\mathbf{X}_k=\mathbf{U}_k\Sigma_k\mathbf{V}_k^T\in \mathbb{R}^{d\times n}$. $\mathbf{U}_k$ and $\mathbf{V}_k$ are the left- and right-singular matrices corresponding to the $k$ biggest singular values of $\mathbf{X}_k$, and $\Sigma_k$ is a diagonal $k\times k$ matrix with $k$ biggest singular values on the diagonal.
$\mathbf{X}_k$ is the best rank-$k$ approximation to $\mathbf{X}$, i.e.,
\begin{align}
\|\mathbf{X} - \mathbf{X}_k\|_F^2 = \min_{\mathbf{Y}\in \mathbb{R}^{m \times n}: \text{rank}(\mathbf{Y})\leq k} \| \mathbf{X} - \mathbf{Y}\|_F^2
\end{align}

\begin{figure}
\centering
{\includegraphics[width=0.85\linewidth]{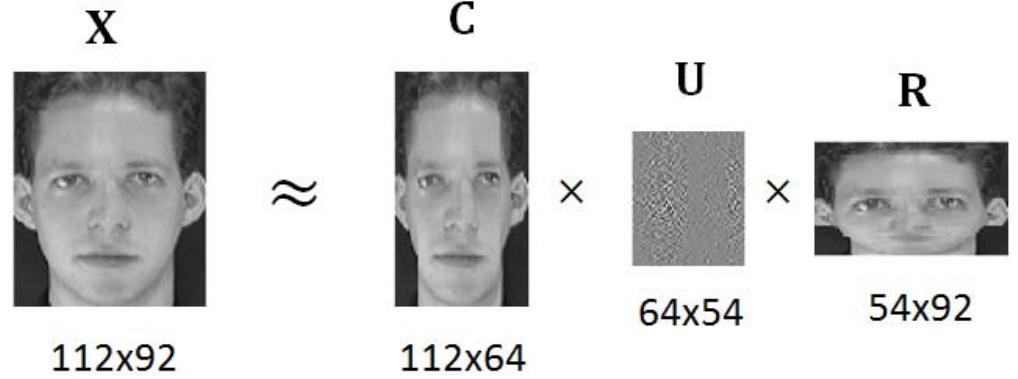}}
\caption{CUR matrix factorization. Image $\mathbf{X}$ comes from the ORL dataset \cite{Samaria1994Harter}. Image $\mathbf{C}$, $\mathbf{U}$, and $\mathbf{R}$ are obtained by \cite{mahoney2009cur}.}
\label{demo}
\end{figure}

In order to further illustrate CUR decomposition, we take Fig. \ref{demo} as an example. $\mathbf{X}$ is a face image. CUR decomposition aims to find image $\mathbf{C}$, $\mathbf{U}$, and $\mathbf{R}$ to reconstruct $\mathbf{X}$ within relative-error ($\mathbf{C}$ is a subset of columns of $\mathbf{X}$ and $\mathbf{R}$ is a  subset of  rows of $\mathbf{X}$.).

From an algorithmic perspective, the matrices $\mathbf{C}$, $\mathbf{U}$, and $\mathbf{R}$ can be obtained by minimizing the approximation error $\|\mathbf{X}-\mathbf{C}\mathbf{U}\mathbf{R}\|_F^2$. Here we make a key observation that the above definition is closely related to the problem of simultaneous sample and feature selection, though to our surprise, existing works rarely point out or explore this connection to solve the active learning problem.

More specifically, the matrix $\mathbf{UR}$ can be regarded as a reconstruction coefficient matrix, and $\mathbf{C}$ denotes the selected $m$ samples, thus minimizing $\|\mathbf{X}-\mathbf{C}\mathbf{U}\mathbf{R}\|_F^2$ means that the total reconstruction error is minimized, which can make the data points in $\mathbf{C}$ be the most representative. The reconstruction coefficients $\mathbf{UR}$ are related to an $r$-dimensional feature subset of the dataset. The reconstruction coefficients of each reconstructed data point $\mathbf{x}_i$ are formed by a linear combination of its $r$ features. In the meantime, the matrix $\mathbf{CU}$ can also be regarded as a reconstruction coefficient matrix, and $\mathbf{R}$ is the new low-dimensional representation of $\mathbf{X}$, so minimizing $\|\mathbf{X}-\mathbf{C}\mathbf{U}\mathbf{R}\|_F^2$ also indicates that the selected $r$ features can represent the whole dataset most precisely. The construction of the coefficient matrix $\mathbf{CU}$ depends on a sample subset of $\mathbf{X}$.
Clearly, active learning and feature selection can be conducted simultaneously in such a joint framework via CUR factorization.

Let $\mathbf{p}=(p_1,\ldots,p_n)^T\in \{0,1\}^n$ and $\mathbf{q}=(q_1,\ldots,q_d)^T\in \{0,1\}^d$ denote two indicator variables to represent whether a sample and a feature is selected or not, respectively. Specifically, $p_i=1$ (or 0) indicates that the $i$-th sample is selected (or not), and $q_i=1$ (or 0) means that the $i$-th feature is selected (or not). Then, minimizing $\|\mathbf{X}-\mathbf{C}\mathbf{U}\mathbf{R}\|_F^2$ can be re-written as
\begin{align}\label{obj1}
\min\limits_{\mathbf{p},\mathbf{q},\widehat{\mathbf{U}}\in \mathbb{R}^{n\times d}}\ &\|\mathbf{X}-\mathbf{X}\text{diag}(\mathbf{p})\widehat{\mathbf{U}}\text{diag}(\mathbf{q})\mathbf{X}\|_F^2\nonumber\\
s.t. \ &\mathbf{1}_n^T\mathbf{p}=m, \mathbf{p}\in \{0,1\}^n,\\
&\mathbf{1}_d^T\mathbf{q}=r, \mathbf{q}\in \{0,1\}^r,\nonumber
\end{align}
where $\text{diag}(\mathbf{p})$ is a diagonal matrix with $\mathbf{p}$ on its diagonal, and $\mathbf{1}_n$ is an $n$-dimensional vector with all components being 1. The term $\mathbf{X}\text{diag}(\mathbf{p})$ in (\ref{obj1}) aims to make $m$ columns of $\mathbf{X}$ unchanged, and resets the rest $n-m$ columns to zero vectors. While $\text{diag}(\mathbf{q})\mathbf{X}$ tends to keep $r$ rows of $\mathbf{X}$ unchanged, and resets the rest $(d-r)$ rows to zero vectors.

Despite the above connection from CUR decomposition to feature selection and active learning, the original CUR formulation and its existing solvers can not be directly applied to solve the simultaneous feature and sample selection task. This is inherently due to the under-determination of a general CUR model. To mitigate this issue, we devise a tailored objective function by adding regularization terms to incorporate prior knowledge.

Moreover, unlike most existing CUR solvers working in a randomized or heuristic fashion \cite{mahoney2009cur,boutsidis2014optimal}, we utilize the structured sparsity-inducing norms to relax the objective from a non-convex optimization problem to a convex one, which allows devising an efficient variant of the alternating direction method of multipliers (ADMM) \cite{gabay1976dual,lin2011linearized}.

\subsection{A Convex Formulation}
The objective function (\ref{obj1}) is hard to be solved directly, since it is an NP-hard problem. After a careful observation to (\ref{obj1}), we find that we can utilize the matrix $l_{2,0}$ norm to reduce the number of the parameters. Defining $\mathbf{W}=\text{diag}(\mathbf{p})\widehat{\mathbf{U}}\text{diag}(\mathbf{q})\in\mathbb{R}^{n \times d}$, we can rewrite (\ref{obj1}) as
\begin{align}\label{obj2}
&\min\limits_{\mathbf{W}\in \mathbb{R}^{n\times d}} \ \|\mathbf{X}-\mathbf{XWX}\|_F^2\nonumber\\
s.t. \ &\|\mathbf{W}\|_{2,0}=m, \|\mathbf{W}^T\|_{2,0}=r,
\end{align}

Based on (\ref{obj2}), we propose to optimize the following objective function:
\begin{align}\label{obj3}
\min\limits_{\mathbf{W}\in \mathbb{R}^{n\times d}} \ &\|\mathbf{X}-\mathbf{XWX}\|_F^2+\alpha\|\mathbf{W}\|_{2,0}+\beta\|\mathbf{W}^T\|_{2,0},
\end{align}
where $\alpha\geq 0$ and $\beta\geq 0$ are two regularization parameters.

However, (\ref{obj3}) is still an NP-hard problem due to the matrix $l_{2,0}$ norm.
Fortunately, there exists theoretical progress that $\|\mathbf{W}\|_{2,1}$ is the minimum convex hull of $\|\mathbf{W}\|_{2,0}$ \cite{nie2013early}.
The result of minimizing $\|\mathbf{W}\|_{2,1}$ is the same as that of minimizing $\|\mathbf{W}\|_{2,0}$, as long as $\mathbf{W}$ is row-sparse enough.
Therefore, (\ref{obj3}) can be relaxed to the following convex optimization problem:
\begin{align}\label{obj4}
\min\limits_{\mathbf{W}\in \mathbb{R}^{n\times d}} \ &\|\mathbf{X}-\mathbf{XWX}\|_F^2+\alpha\|\mathbf{W}\|_{2,1}+\beta\|\mathbf{W}^T\|_{2,1}.
\end{align}

\subsection{Local Linear Reconstruction}
In the new objective function (\ref{obj4}), each data point is reconstructed by a linear combination of all selected points (when the $i$-th row of the reconstruction coefficient matrix $\mathbf{WX}$ in (\ref{obj4}) is not a zero vector, $\mathbf{x}_i$ is chosen as one of the most representative samples. Otherwise, $\mathbf{x}_i$ is not selected).
However, it is more reasonable to suppose that a data point can be mainly recovered from its neighbors \cite{hu2013active,cai2012manifold}.
Intuitively, if the distance between the reconstructed point and the selected point is large, the contribution of the selected point should be small to the reconstruction of the target point, and thus the reconstruction coefficient should be penalized. In light of this point, we incorporate a regularization term into (\ref{obj4}) as
\begin{align}\label{obj5}
\min\limits_{\mathbf{W}\in \mathbb{R}^{n\times d}} \ &\|\mathbf{X}-\mathbf{XWX}\|_F^2+\alpha\|\mathbf{W}\|_{2,1}+\beta\|\mathbf{W}^T\|_{2,1}\nonumber\\
&+\lambda \|\mathbf{T}\odot(\mathbf{WX})\|_1,
\end{align}
where $\lambda \geq 0$ is a regularization parameter, and $\odot$ denotes the element-wise multiplication of two matrices. $\mathbf{T}$ is a weight matrix, where $\mathbf{T}_{ij}$ encodes the distance between the $i$-th and $j$-th samples. From the data reconstruction perspective, if two unit vectors have the same or opposite directions, their distance should be minimal, since either vector can be fully recovered by the other one; on the contrary, if the two vectors are orthogonal, their distance should be maximal, because they have little contribution to each other's reconstruction. Therefore, we use the absolute value of the cosine function of the angle between two feature vectors to measure their similarity, and define the inverse of the absolute value as their distance:
\begin{align}
\mathbf{T}_{ij}=\frac{1}{|\cos \theta_{ij}|},
\end{align}
where $\theta_{ij}$ denotes the angle between $\mathbf{x}_i$ and $\mathbf{x}_j$\footnote{When $\cos\theta_{ij}=0$, we can regularize $\mathbf{T}_{ij}$ as $\mathbf{T}_{ij}=\frac{1}{|\cos \theta_{ij}|+\varsigma}$, where $\varsigma$ is a very small positive constant.}.

After obtaining the optimal $\mathbf{W}$ in (\ref{obj5}), we can sort all the samples by the $l_2$ norm of the rows of $\mathbf{W}$ in descending order, and select the top $m$ samples as the representative ones. Similarly, we rank all the features by the $l_2$ norm of the columns of $\mathbf{W}$ in descending order, and choose the top $r$ features to represent the samples.

\begin{figure}
\centering
\subfigure[$\|\mathbf{W}\|_{2,1}$]{\includegraphics[width=0.4\linewidth]{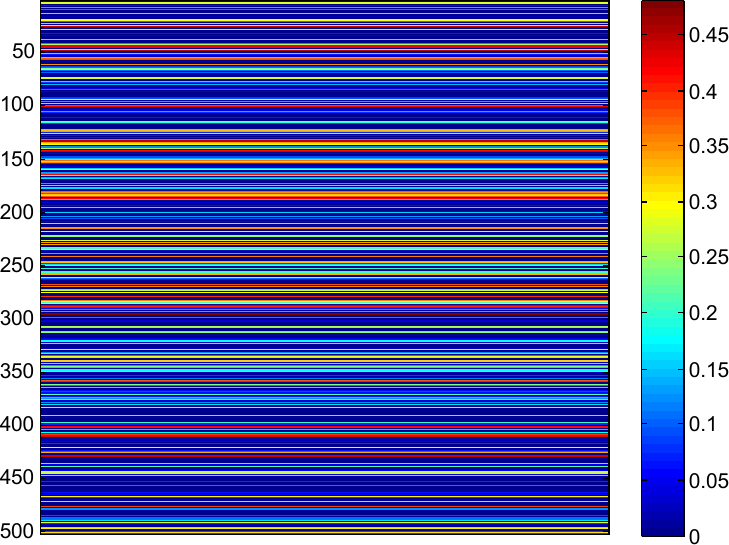}}
\subfigure[$\|\mathbf{W}^T\|_{2,1}$]{\includegraphics[width=0.38\linewidth]{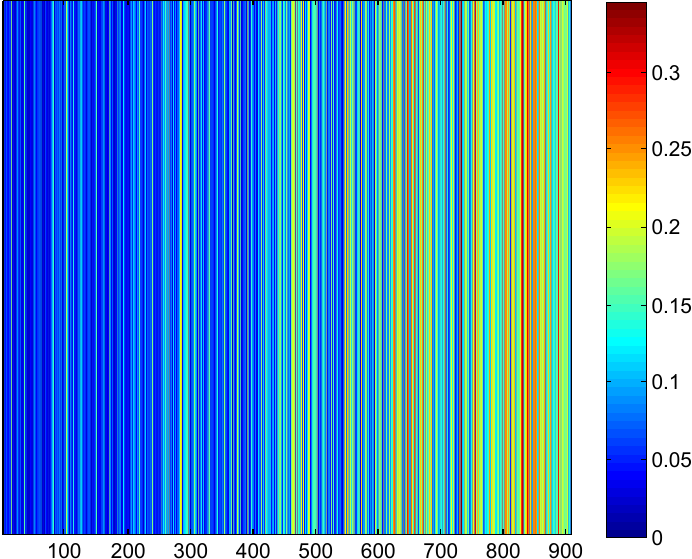}}
\vspace{-0.1in}
\caption{The visualization of the learned $\mathbf{W}$ on the FG-NET dataset. (a) Each row is the $l_2$ norm value of each row of $\mathbf{W}$. (b) Each column is the $l_2$ norm value of each column of $\mathbf{W}$. Dark blue denotes that the values are close to zero.}
\label{visual_norm}
\end{figure}

 We take the FG-NET dataset\footnote{The dataset is available at \url{http://sting.cycollege.ac.cy/alanitis/fgnetaging/index.htm}.} as an example to illustrate the effectiveness of the $l_{2,1}$ norm constraint on $\mathbf{W}$ and $\mathbf{W}^T$ in (\ref{obj5}). Fig. \ref{visual_norm} (a) and (b) are the visualizations of $l_{2,1}$ norm of $\mathbf{W}$ and $\mathbf{W}^T$, respectively. Many rows and columns in $\mathbf{W}$ become sparse by adding the $l_{2,1}$ norm constraints on $\mathbf{W}$ and $\mathbf{W}^T$, which means that $\mathbf{W}$ can conduct sample selection and feature selection simultaneously.

\begin{figure}
\centering
{\includegraphics[width=0.6\linewidth]{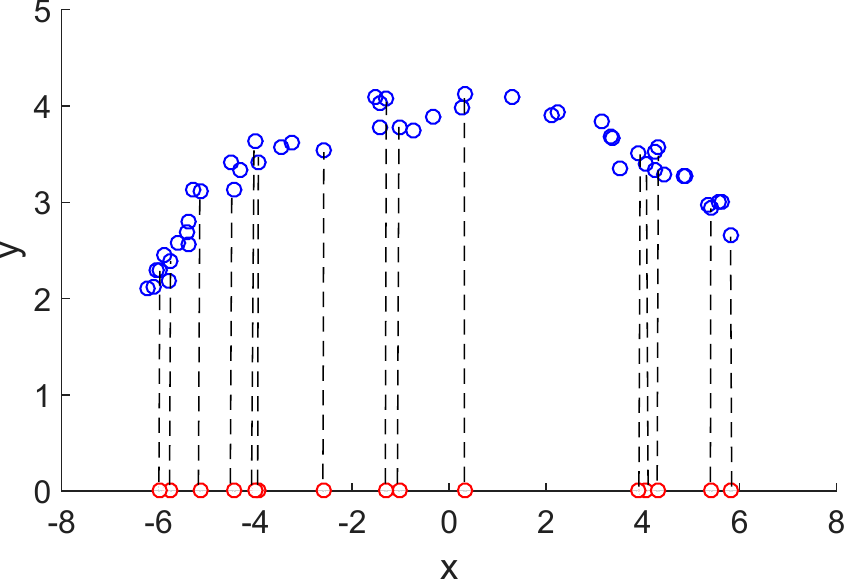}}
\vspace{-0.1in}
\caption{Data and feature selection by our method on a synthetic dataset. Blue circles denote original 2-D data points, and red circles denote the 1-D data points selected by our method.}
\label{toy}
\end{figure}

 We also apply our method on a synthetic dataset to give an intuitive idea of how our method works. In Fig. \ref{toy}, blue circles denote original 2-D data points. The feature in the direction of the x-axis has more information (in terms of variance) than one in the y-axis direction. Red circles denote the 1-D data points selected by our method. Based on the results, our method can select representative samples and features.

\subsection{Optimization Algorithm}
Although the problem (\ref{obj5}) is convex, it is not easy to be solved by sub-gradient type methods since different structured non-smooth terms are involved.
In \cite{nie2013early}, the authors proposed an easier algorithm by letting the derivative of the objective function be zero directly. But this technique does not fit our model, because our objective function has three non-smooth terms, {\em{i.e.}}, $\alpha \left\| \mathbf{W} \right\|_{2,1}$, $\beta \left\| \mathbf{W}^T \right\|_{2,1}$ and $\lambda \left\| \mathbf{T}\odot \mathbf{(WX)} \right\|_{1}$. As a result, it is hard to guarantee that there is an easier solution by taking the derivative of the proposed objective function directly.
In this section, we employ the alternating direction method of multipliers (ADMM) \cite{gabay1976dual} to solve (\ref{obj5}).
The advantage of ADMM is that it can separate the joint problem with respect to three difficult terms $\alpha \left\| \mathbf{W} \right\|_{2,1}$, $\beta \left\| \mathbf{W}^T \right\|_{2,1}$ and $\lambda \left\| \mathbf{T}\odot \mathbf{(WX)} \right\|_{1}$ into three easier sub-problems. Then, the resulting three sub-problems are much easier to calculate and all of them have closed-from solutions.

In order to solve (\ref{obj5}), we first introduce three variables $\widehat{\mathbf{W}}$, $\widetilde{\mathbf{W}}$ and $\mathbf{Z}$, to convert (\ref{obj5}) to the following equivalent objective function:
\begin{align}\label{obj6}
\min\limits_{\mathbf{W}, \widehat{\mathbf{W}}, \widetilde{\mathbf{W}}, \mathbf{Z}} \ &\|\mathbf{X}-\mathbf{XWX}\|_F^2+\alpha\|\widehat{\mathbf{W}}\|_{2,1}+\beta\|\widetilde{\mathbf{W}}\|_{2,1}\nonumber\\
&+\lambda \|\mathbf{T}\odot \mathbf{Z}\|_1 \nonumber\\
s.t. \ &\mathbf{WX}=\mathbf{Z}, \mathbf{W}=\widehat{\mathbf{W}}, \mathbf{W}^T=\widetilde{\mathbf{W}}.
\end{align}
The augmented Lagrange function of (\ref{obj6}) is
\begin{align}\label{obj7}
&\  \mathcal{L}_{\rho_1,\rho_2, \rho_3}(\mathbf{W},\widehat{\mathbf{W}}, \widetilde{\mathbf{W}}, \mathbf{Z}, \Lambda_1, \Lambda_2, \Lambda_3):=\|\mathbf{X}-\mathbf{XWX}\|_F^2\nonumber\\
&+\alpha\|\widehat{\mathbf{W}}\|_{2,1}+\beta\|\widetilde{\mathbf{W}}\|_{2,1}+\lambda \|\mathbf{T}\odot \mathbf{Z}\|_1 +\langle\Lambda_1,\mathbf{WX}-\mathbf{Z}\rangle\nonumber\\
&+\langle\Lambda_2,\mathbf{W}-\widehat{\mathbf{W}}\rangle +\langle\Lambda_3,\mathbf{W}^T-\widetilde{\mathbf{W}}\rangle +\frac{\rho_1}{2}\|\mathbf{WX}-\mathbf{Z}\|_F^2\nonumber\\
&+\frac{\rho_2}{2}\|\mathbf{W}-\widehat{\mathbf{W}}\|_F^2+\frac{\rho_3}{2}\|\mathbf{W}^T-\widetilde{\mathbf{W}}\|_F^2,
\end{align}
where $\Lambda_1$, $\Lambda_2$, $\Lambda_3$ are Lagrange multipliers. $\rho_1$, $\rho_2$, $\rho_3$ are the constraint violation penalty parameters.
From the augmented Lagrangian function, we can find that the subproblems about $\widehat{\mathbf{W}}$, $\widetilde{\mathbf{W}}$ and $\mathbf{Z}$ are fully separable, as a result we can introduce the classical two-block ADMM here, while considering $\mathbf{W}$ and $(\widehat{\mathbf{W}}, \widetilde{\mathbf{W}}, \mathbf{Z})$ as two-block variables. Recall that in a ADMM-type algorithm, the basic Gauss-Seidel structure in ($t+1$)-th iteration is as
\begin{equation}
\!\!\!\!\left\{\!\!
\begin{array}{l}
\mathbf{W}^{k+1} = \arg\min\mathcal{L} (\mathbf{W},\widehat{\mathbf{W}}^k,\widetilde{\mathbf{W}}^k,\mathbf{Z}^k, \Lambda_1^k, \Lambda_2^k,\Lambda_3^k),\\
\widehat{\mathbf{W}}^{k+1} =  \arg\min\mathcal{L} (\mathbf{W}^{k+1},\widehat{\mathbf{W}},\widetilde{\mathbf{W}}^k, \mathbf{Z}^k, \Lambda_1^k, \Lambda_2^k,\Lambda_3^k),\\
\widetilde{\mathbf{W}}^{k+1} =  \arg\min\mathcal{L} (\mathbf{W}^{k+1},\widehat{\mathbf{W}}^{k+1},\widetilde{\mathbf{W}}, \mathbf{Z}^k, \Lambda_1^k, \Lambda_2^k,\Lambda_3^k),\\
\mathbf{Z}^{k+1} =  \arg\min\mathcal{L} (\mathbf{W}^{k+1},\widehat{\mathbf{W}}^{k+1},\widetilde{\mathbf{W}}^{k+1}, \mathbf{Z}, \Lambda_1^k, \Lambda_2^k,\Lambda_3^k),\\
\Lambda_1^{k+1} = \Lambda_1^k + \rho_1 ( \mathbf{W}^{k+1}\mathbf{X} - {\mathbf{Z}}^{k+1} ),\\
\Lambda_2^{k+1} = \Lambda_2^k + \rho_2 ( \mathbf{W}^{k+1} - \widehat{\mathbf{W}}^{k+1} ),\\
\Lambda_3^{k+1} = \Lambda_3^k + \rho_3 ( (\mathbf{W}^{k+1})^T - \widetilde{\mathbf{W}}^{k+1} ). \nonumber
\end{array}
\right.
\end{equation}

Next, we will introduce how to solve these subproblems in detail.

i) Compute the subproblem about $\mathbf{W}^{k+1}$: When the other variables are fixed with the former iteration result $(\widehat{\mathbf{W}}^k, \widetilde{\mathbf{W}}^k, \mathbf{Z}^{k}, \Lambda_1^k, \Lambda_2^k, \Lambda_3^k)$, the subproblem about $\mathbf{W}^{k+1}$ is as
\begin{align}
\mathbf{W}&^{k+1}=\arg\min_{\mathbf{W}}\mathcal{L}_{\rho_1,\rho_2,\rho_3}(\mathbf{W},\widehat{\mathbf{W}}^k,\widetilde{\mathbf{W}}^k,\mathbf{Z}^k, \Lambda_1^k, \Lambda_2^k,\Lambda_3^k)\nonumber\\
=&\arg \min_{\mathbf{W}}\|\mathbf{X}-\mathbf{XWX}\|_F^2+\frac{\rho_1}{2}\|\mathbf{WX}-\mathbf{Z}^k+\frac{\Lambda_1^k}{\rho_1}\|_F^2 \nonumber\\
& +\frac{\rho_2}{2}\|\mathbf{W}-\widehat{\mathbf{W}}^k+\frac{\Lambda_2^k}{\rho_2}\|_F^2
+\frac{\rho_3}{2}\|\mathbf{W}^T-\widetilde{\mathbf{W}}^k+\frac{\Lambda_3^k}{\rho_3}\|_F^2.\nonumber
\end{align}

The necessary optimality condition further follows as
\begin{align}
\frac{\partial\mathcal{L}_{\rho_1,\rho_2,\rho_3}(\mathbf{W},\widehat{\mathbf{W}}^k,\widetilde{\mathbf{W}}^k,\mathbf{Z}^k, \Lambda_1^k, \Lambda_2^k,\Lambda_3^k) }{\partial \mathbf{W}}=0.
\end{align}

This implies
\begin{align}
(2\mathbf{X}^T\mathbf{X} + \rho_1 \mathbf{I})\mathbf{W} \mathbf{X}\mathbf{X}^T+(\rho_2+\rho_3)\mathbf{W} =2\mathbf{X}^T\mathbf{X}\mathbf{X}^T \nonumber\\
+ \rho_1(\mathbf{Z}^k-\frac{\Lambda_1^k}{\rho_1})\mathbf{X}^T +\rho_2(\widehat{\mathbf{W}}^k-\frac{\Lambda_2^k}{\rho_2})
+\rho_3(\widetilde{\mathbf{W}}^k-\frac{\Lambda_3^k}{\rho_3})^T. \nonumber
\end{align}

For writing conveniently, let $\mathbf{M}=2\mathbf{X}^T\mathbf{X} + \rho_1 \mathbf{I}$, and $\mathbf{H}=2\mathbf{X}^T\mathbf{X}\mathbf{X}^T
+ \rho_1(\mathbf{Z}^k-\frac{\Lambda_1^k}{\rho_1})\mathbf{X}^T +\rho_2(\widehat{\mathbf{W}}^k-\frac{\Lambda_2^k}{\rho_2})
+\rho_3(\widetilde{\mathbf{W}}^k-\frac{\Lambda_3^k}{\rho_3})^T$, then the equation above becomes
\begin{align}\label{w2}
\mathbf{M}\mathbf{W} \mathbf{X}\mathbf{X}^T+(\rho_2+\rho_3)\mathbf{W}=\mathbf{H}.
\end{align}

Since $\mathbf{M}$ and $\mathbf{X}\mathbf{X}^T$  are positive semi-definite and symmetric, we can perform eigenvalue decomposition with all non-negative eigenvalues, obtaining
\begin{equation}\label{decom11}
\!\!\!\!\left\{\!\!
\begin{array}{l}
\mathbf{M}=\mathbf{P}\Theta_1\mathbf{P}^T,\\
\mathbf{X}\mathbf{X}^T=\mathbf{Q}\Theta_2\mathbf{Q}^T,\qquad \qquad \qquad \qquad \qquad \qquad
\end{array}
\right.
\end{equation}
where $\mathbf{P}$ and $\mathbf{Q}$ are both orthogonal. $\Theta_1$ and $\Theta_2$ are two diagonal matrices.

Plugging (\ref{decom11}) into (\ref{w2}), we obtain
\begin{align}\label{w3}
&\mathbf{P}\Theta_1\mathbf{P}^T\mathbf{W}\mathbf{Q}\Theta_2\mathbf{Q}^T+(\rho_2+\rho_3)\mathbf{W}=\mathbf{H}\nonumber\\
\Rightarrow &\Theta_1\mathbf{P}^T\mathbf{W}\mathbf{Q}\Theta_2 +(\rho_2+\rho_3) \mathbf{P}^T \mathbf{W}\mathbf{Q}=\mathbf{P}^T \mathbf{H}\mathbf{Q}.
\end{align}

Let $\mathbf{Y}=\mathbf{P}^T\mathbf{W}\mathbf{Q}$, then (\ref{w3}) becomes
\begin{align}
&\ \ \ \ \ \ \ \ \Theta_1\mathbf{Y}\Theta_2+(\rho_2+\rho_3)\mathbf{Y}=\mathbf{P}^T \mathbf{H}\mathbf{Q}\nonumber\\
&\Rightarrow \mathbf{Y}_{ij}=\frac{(\mathbf{P}^T \mathbf{H}\mathbf{Q})_{ij}}{(\Theta_1)_{ii}(\Theta_2)_{jj}\!+\!\rho_2\!+\!\rho_3}, i=1,\ldots,n, j=1,\ldots, d.\nonumber
\end{align}

As we know, $(\Theta_1)_{ii}\geq0, (\Theta_2)_{jj}\geq0$. In the meantime, $\rho_2$ and $\rho_3$ are greater than zero in practice, so the denominator in the equation above is greater than zero. After obtaining $\mathbf{Y}$, we can easily calculate $\mathbf{W}^{k+1}$ as
\begin{align}\label{optimalw}
\mathbf{W}^{k+1}=\mathbf{PYQ}^T
\end{align}

ii) Further we calculate the subproblem about $\widehat{\mathbf{W}}^{k+1}$, i.e.,
\begin{align}\label{what}
\widehat{\mathbf{W}}^{k+1}=\arg\min_{\widehat{\mathbf{W}}} \mathcal{L} (\mathbf{W}^{k+1},\widehat{\mathbf{W}},\widetilde{\mathbf{W}}^k, \mathbf{Z}^k, \Lambda_1^k, \Lambda_2^k,\Lambda_3^k)\nonumber\\
=\arg\min_{\widehat{\mathbf{W}}} \alpha\|\widehat{\mathbf{W}}\|_{2,1} +\frac{\rho_2}{2}\|\widehat{\mathbf{W}}-\mathbf{W}^{k+1}-\frac{\Lambda_2^k}{\rho_2}\|_F^2.
\end{align}

In order to solve the subproblem (\ref{what}), we first decouple it as
\begin{align}\label{what1}
\widehat{\mathbf{W}}^{k+1}=&\arg\min_{\widehat{\mathbf{W}}^i}\sum_{i=1}^n \alpha\|\widehat{\mathbf{W}}^i\|_{2} \nonumber\\ &+\frac{\rho_2}{2}\sum_{i=1}^n\|\widehat{\mathbf{W}}^i-(\mathbf{W}^{k+1}+\frac{\Lambda_2^k}{\rho_2})^i\|_2^2,
\end{align}
where ${\widehat{\mathbf{W}}^i}$ and $(\mathbf{W}^{k+1}+\frac{1}{{\rho_2}}{\Lambda_2^k})^i$ are the $i$-th row of matrix $\widehat{\mathbf{W}}$ and $\mathbf{W}^{k+1}+\frac{1}{{\rho_2}}{\Lambda_2^k}$ respectively. The problem (\ref{what1}) can be solved by the following lemma \cite{yang2009fast}:
\newtheorem{thm}{Lemma}[section]
\begin{thm}\label{theo1}
For any $\sigma, \eta >0$, and $\mathbf{v}\in \mathbb{R}^{q}$, the minimizer of
\begin{align}
\min_{\mathbf{u}\in \mathbb{R}^{q}}\sigma\|\mathbf{u}\|_2+\frac{\eta}{2}\|\mathbf{u}-\mathbf{v}\|_2^2,
\end{align}
is given by
\begin{equation}\label{decom}
\mathbf{u}=\left\{
\begin{array}{l}
(1-\frac{\sigma}{\eta \|\mathbf{v}\|_2})\mathbf{v}, \ \ \ \ \|\mathbf{v}\|_2> \frac{\sigma}{\eta}\\
\ \ \ \ \ \ \ \ 0, \ \ \ \ \ \  \ \ \ \ \ \ \ \ \ \|\mathbf{v}\|_2\leq \frac{\sigma}{\eta}.
\end{array}
\right.
\end{equation}
\end{thm}

Based on this lemma, we can obtain the optimal $\widehat{\mathbf{W}}^{k+1}$ as
\begin{equation}\label{optimalwhat}
(\widehat{\mathbf{W}}^{k+1})^i=\left\{
\begin{array}{l}
(1-\frac{\alpha}{\rho_2 \|\mathbf{s}\|_2})\mathbf{s}, \ \ \ \ \|\mathbf{s}\|_2> \frac{\alpha}{\rho_2}\\
\ \ \ \ \ \ \ \ 0, \ \ \ \ \ \  \ \ \ \ \ \ \ \ \ \|\mathbf{s}\|_2\leq \frac{\alpha}{\rho_2},
\end{array}
\right.
\end{equation}
where $\mathbf{s}=(\mathbf{W}^{k+1}+\frac{1}{{\rho_2}}{\Lambda_2^k})^i$.

iii) $\widetilde{\mathbf{W}}^{k+1}$ is the minimizer for
\begin{align}
\min_{\widetilde{\mathbf{W}}}\mathcal{L} (\mathbf{W}^{k+1},\widehat{\mathbf{W}}^{k+1},\widetilde{\mathbf{W}}, \mathbf{Z}^k, \Lambda_1^k, \Lambda_2^k,\Lambda_3^k) \nonumber\\
=\min_{\widetilde{\mathbf{W}}}\beta\|\widetilde{\mathbf{W}}\|_{2,1} +\frac{\rho_3}{2}\|\widetilde{\mathbf{W}}-\left((\mathbf{W}^{k+1})^T+\frac{\Lambda_3^k}{\rho_3}\right)\|_F^2
\end{align}

Similar to solve (\ref{what}), the optimal $\widetilde{\mathbf{W}}^{k+1}$ can be easily obtained by
\begin{equation}\label{optimalwtilde}
(\widetilde{\mathbf{W}}^{k+1})^i=\left\{
\begin{array}{l}
(1-\frac{\beta}{\rho_3 \|\mathbf{s}\|_2})\mathbf{s}, \ \ \ \ \|\mathbf{s}\|_2> \frac{\beta}{\rho_3}\\
\ \ \ \ \ \ \ \ 0, \ \ \ \ \ \  \ \ \ \ \ \ \ \ \ \|\mathbf{s}\|_2\leq \frac{\beta}{\rho_3},
\end{array}
\right.
\end{equation}
where $\mathbf{s}=\left((\mathbf{W}^{k+1})^T+\frac{1}{{\rho_3}}{\Lambda_3^k}\right)^i$.

iv) In order to compute the subproblem about $\mathbf{Z}^{k+1}$, we need to solve
\begin{align}\label{subz}
\min_{{\mathbf{Z}}}(\mathbf{W}^{k+1},\widehat{\mathbf{W}}^{k+1},\widetilde{\mathbf{W}}^{k+1}, \mathbf{Z}, \Lambda_1^k, \Lambda_2^k,\Lambda_3^k)\nonumber\\
=\min_{{\mathbf{Z}}} \lambda \|\mathbf{T}\odot \mathbf{Z}\|_1+\frac{\rho_1}{2}\|{\mathbf{Z}}-\mathbf{W}^{k+1}\mathbf{X}-\frac{\Lambda_1^k}{\rho_1}\|_F^2.
\end{align}

The problem (\ref{subz}) can be solved by the following matrix shrinkage operation Lemma \cite{lin2009augmented}:

\begin{thm}\label{theo1}
For $\mu >0$, and $\mathbf{K}\in \mathbb{R}^{s\times t}$, the solution of the problem
\begin{align}
\min_{\mathbf{L}\in \mathbb{R}^{s\times t}}\mu\|\mathbf{L}\|_1+\frac{1}{2}\|\mathbf{L}-\mathbf{K}\|_F^2, \nonumber
\end{align}
is given by $\mathnormal{L}_\mu(\mathbf{K})\in\mathbb{R}^{s\times t}$, which is defined componentwisely by
\begin{align}\label{lmuk}
(\mathnormal{L}_\mu(\mathbf{K}))_{ij}:=\max\{|\mathbf{K}_{ij}|-\mu, 0\}\cdot sgn(\mathbf{K}_{ij}),
\end{align}
\end{thm}
where $sgn(t)$ is the signum function of $t\in \mathbf{R}$, i.e.,
\begin{align}
sgn(t):=
\left\{
\begin{array}{cc}
+1 & \text{if} \ t>0, \nonumber\\
0 & \text{if} \ t=0, \nonumber\\
-1 & \text{if}\  t<0.\nonumber
\end{array}
\right.
\end{align}

Based on Lemma \ref{theo1}, we can obtain a closed-form solution of $\mathbf{Z}^{k+1}$ whose ($i,j$)-th entry is expressed as
\begin{align}\label{optZ}
\mathbf{Z}_{ij}^{k+1}:=&\max\{|(\mathbf{W}^{k+1}\mathbf{X}+\frac{\Lambda_1^k}{\rho})_{ij}|-\frac{\lambda\cdot\mathbf{T}_{ij}}{\rho_1},0\}\nonumber\\
&\cdot sgn((\mathbf{W}^{k+1}\mathbf{X}+\frac{\Lambda_1^k}{\rho_1})_{ij}).
\end{align}

The key steps of the proposed ALFS algorithm are summarized in Algorithm 1.
We can also extend our method to the kernel version by defining a new data representation to incorporate the kernel information as in \cite{zhang2013learning}.

\begin{table}
\begin{center}
\label{activefeaturelearning}
\begin{tabular}{l}
\hline
\textbf{Algorithm 1} \   The ALFS Algorithm \\
\hline
\textbf{Input:} The data matrix $\mathbf{X}\in\mathbb{R}^{d\times n}$, parameters $\alpha$, $\beta$, and $\lambda$.\\
 \textbf{Initialize}: $\mathbf{W}^0=\widehat{{\mathbf{W}}}^0=\textbf{0}$, $\widetilde{\mathbf{W}}^\textbf{0}=0$, $\mathbf{Z}^0=\textbf{0}$, $\Lambda_1^0=\textbf{0}$, $\Lambda_2^0=\textbf{0}$,  $\Lambda_3^0=\textbf{0}$, \\
 $\rho_1=\rho_2=\rho_3=10^{-6}, \max_\rho=10^{10}$, $\tau=1.1$, $\epsilon=10^{-3}$, $k=0$.\\
\textbf{while} not converged \textbf{do} \\
 \ \ \ \ \ \ 1. fix the other variables and update $\mathbf{W}^{k+1}$ by (\ref{optimalw});\\
 \ \ \ \ \ \ 2. fix the other variables and update $\widehat{\mathbf{W}}^{k+1}$ by (\ref{optimalwhat});\\
 \ \ \ \ \ \ 3. fix the other variables and update $\widetilde{\mathbf{W}}^{k+1}$ by (\ref{optimalwtilde});\\
  \ \ \ \ \ \ 4. fix the other variables and update ${\mathbf{Z}}^{k+1}$ by (\ref{optZ});\\
 \ \ \ \ \ \ 5. update the multipliers\\
  \ \ \ \ \ \ \ \ \ \ $\Lambda_1^{k+1}=\Lambda_1^{k}+\rho_1(\mathbf{W}^{k+1}\mathbf{X}-\mathbf{Z}^{k+1})$,\\
  \ \ \ \ \ \ \ \ \ \ $\Lambda_2^{k+1}=\Lambda_2^{k}+\rho_2(\mathbf{W}^{k+1}-\widehat{\mathbf{W}}^{k+1})$,\\
    \ \ \ \ \ \ \ \ \ \ $\Lambda_3^{k+1}=\Lambda_3^{k}+\rho_3\left((\mathbf{W}^{k+1})^T-\widetilde{\mathbf{W}}^{k+1}\right)$;\\
 \ \ \ \ \ \ 6. update the parameters $\rho_1$, $\rho_2$, and $\rho_3$ by \\
  \ \ \ \ \ \  \ \ \ \ \ $\rho_i = \min(\tau\rho_i, \max_\rho)$, i=1,2,3;\\
  \ \ \ \ \ \ 7. $k\leftarrow k+1$;\\
  \ \ \ \ \ \ 8. check the convergence conditions\\
  \ \ \ \ \ \ \ \ \ \ $\|\mathbf{W}^k\mathbf{X}-\mathbf{Z}^k\|_\infty <\epsilon$ and $\|\mathbf{W}^k-\widehat{\mathbf{W}}^k\|_\infty<\epsilon$ and \\
  \ \ \ \ \ \ \ \ \ \ $\|(\mathbf{W}^k)^T-\widetilde{\mathbf{W}}^k\|_\infty<\epsilon$ and $|\frac{f(\mathbf{W}^k)-f(\mathbf{W}^{k-1})}{f(\mathbf{W}^{k-1})}|<\epsilon$, where \\
  \ \ \ \ \ \ \ \ \ \ $f(\mathbf{W}^k)$ is the objective function value of (\ref{obj5}) at the point $\mathbf{W}^k$. \\
 \textbf{end while}\\
 \textbf{Output:} The matrix $\mathbf{W}^k\in\mathbb{R}^{n\times d}$.\\
\hline
\end{tabular}
\end{center}
\end{table}

\subsection{Algorithm Analysis}
{From the framework of ALFS, we can find that Algorithm 1 is the direct application of the classical two-block ADMM, although the problem has more than two block variables. All the subproblems in Algorithm 1 have closed-form solutions. Based on the classical convergence results, we can obtain the global convergence of {{Algorithm 1}} to the primal-dual optimal solution of problem (\ref{obj6}) (see \cite{boyd2011distributed,HeLiaoHanYang2002}).
In the following we present both the global convergence and the iteration complexity results of Algorithm 1.

\begin{theorem}
For given constant parameters $\alpha$, $\beta$, $\gamma$ and given constant penalty parameters $\rho_1$, $\rho_2$, $\rho_3$. Denote the iteration sequence generated by {{Algorithm 1}} as
\begin{equation}
\!\!\!\!\left\{\!\!
\begin{array}{l}
 {\mathbf{\Sigma}}^k := \left\{{\mathbf{W}}^k,{\mathbf{\widehat{W}}}^k,{\mathbf{\widetilde{W}}}^k,{\mathbf{Z}}^k,\Lambda_1^k,\Lambda_2^k,\Lambda_3^k  \right\},\nonumber \qquad \qquad \qquad \\
 {\mathbf{\tilde{\Sigma}}}^k := \frac{1}{k+1} \sum_{t=0}^k {\mathbf{\Sigma}}^t,\nonumber\\
{\mathbf{\Sigma}}_1^k := \left\{{\mathbf{W}}^k,{\mathbf{\widehat{W}}}^k,{\mathbf{\widetilde{W}}}^k,{\mathbf{Z}}^k \right\}, \nonumber\\
{\mathbf{\Sigma}}_2^k := \left\{\Lambda_1^k,\Lambda_2^k,\Lambda_3^k \right\}.\nonumber
\end{array}
\right.
\end{equation}
Then we have the following results:
\begin{enumerate}
\item (Global Convergence) The sequence $\left\{ {\mathbf{\Sigma}}^k \right\}$ converges to a primal-dual optimal solution pair $({\mathbf{W}}^{\infty}$,${\mathbf{\widehat{W}}}^{\infty}$,${\mathbf{\widetilde{W}}}^{\infty}$,${\mathbf{Z}}^{\infty}$,$\Lambda_1^{\infty}$,$\Lambda_2^{\infty}$,$\Lambda_3^{\infty})$, where (${\mathbf{W}}^{\infty}$,${\mathbf{\widehat{W}}}^{\infty}$,${\mathbf{\widetilde{W}}}^{\infty}$,${\mathbf{Z}}^{\infty}$) is the global optimal solution of problem (\ref{obj6})
 and ${\mathbf{W}}^{\infty}$ is the global optimal solution of problem (\ref{obj5}).
\item (Constraint Satisfactory) Both constraint violations will converge to zero, e.g.
\begin{equation}
\!\!\!\!\left\{\!\!
\begin{array}{l}
\left\| {\mathbf{W}}^k {\mathbf{X}} - {\mathbf{Z}}^k \right\|_F \rightarrow 0, \nonumber\qquad \qquad \qquad \qquad \qquad \qquad \qquad \qquad \qquad \\
\left\| {\mathbf{W}}^k  - {\mathbf{\widehat{W}}}^k  \right\|_F \rightarrow 0, \nonumber\\
\left\| ({\mathbf{W}}^k)^T- {\mathbf{\widetilde{W}}}^k  \right\|_F \rightarrow 0.\nonumber
\end{array}
\right.
\end{equation}
\item (Ergodic Iteration Complexity \cite{HeYuan2012}) Let $({\mathbf{W}}^*$,${\mathbf{\widehat{W}}}^*$,${\mathbf{\widetilde{W}}}^*$,${\mathbf{Z}}^*$,$\Lambda_1^*$,$\Lambda_2^*$,$\Lambda_3^*)$ be an optimal solution pair, we have
\begin{equation}\label{Ergodic_IC}
{\cal{L}}_{\rho_1,\rho_2,\rho_3} ({\mathbf{\tilde{\Sigma}}}_1^k, {\mathbf{\Sigma}}_2^*) - {\cal{L}}_{\rho_1,\rho_2,\rho_3} ({\mathbf{\Sigma}}_1^*, {\mathbf{\tilde{\Sigma}}}_2^k) \le \frac{ C_1 }{k+1},
\end{equation}
where $C_1$ denotes a constant related with ${\mathbf{\Sigma}}^0$ and ${\mathbf{\Sigma}}^*$.
\item (Non-ergodic Iteration Complexity \cite{HeYuan2014}) The non-ergodic iteration complexity can be written as
\begin{equation}\label{Nonergodic_IC}
\left\| {\mathbf{\Sigma}}^k - {\mathbf{\Sigma}}^{k+1} \right\|_{\cal{H}}^2 \le \frac{C_2}{k+1},
\end{equation}
where $C_2$ also denotes a constant related with ${\mathbf{\Sigma}}^0$ and ${\mathbf{\Sigma}}^*$ and ${\cal{H}}$ is a matrix related with ${\mathbf{X}}$ as follows,
\begin{equation}
{\cal{H}} = \left( \begin{array}{ccccccc}
				\mathbf{S}&{\mathbf{0}}&{\mathbf{0}}&{\mathbf{0}}&{\mathbf{0}}&{\mathbf{0}}&{\mathbf{0}} \\
				{\mathbf{0}} & \rho_2 {\mathbf{I}} & \ddots & \ddots& \ddots & \ddots&{\mathbf{0}}\\
				{\mathbf{0}} & \ddots & \rho_3 {\mathbf{I}} & \ddots & \ddots & \ddots&{\mathbf{0}}\\
				{\mathbf{0}} & \ddots & \ddots & \rho_1 {\mathbf{I}} & \ddots& \ddots&{\mathbf{0}}\\
				{\mathbf{0}} & \ddots & \ddots& \ddots & \frac{1}{\rho_1} {\mathbf{I}}& \ddots&{\mathbf{0}}\\
				{\mathbf{0}} & \ddots & \ddots& \ddots& \ddots & \frac{1}{\rho_2} {\mathbf{I}}&{\mathbf{0}}\\
				{\mathbf{0}} & {\mathbf{0}} & {\mathbf{0}} & {\mathbf{0}}& {\mathbf{0}} & {\mathbf{0}} & \frac{1}{\rho_3}{\mathbf{I}}\\
				\end{array} \right),\nonumber
\end{equation}
where $\mathbf{S}=\rho_1 {\mathbf{X}}^T {\mathbf{X}} + (\rho_2+\rho_3) {\mathbf{I}}.$
\end{enumerate}
\end{theorem}
For the detailed proof, refer to \cite{HeLiaoHanYang2002,HeYuan2012}. The first and second parts of this theorem show the global convergence of the presented algorithm, including sequence convergence and constraint convergence. From the first part, we can find that the sequence converges to the primal-dual optimal solution pair, while the second part shows the two linear constraints converge to zero in the sense of {\em{Frobenius}} norm.

The third and fourth parts above show a global convergence speed of ADMM, in the sense of ergodic and non-ergodic respectively. The inequality (\ref{Ergodic_IC}) is the ergodic iteration complexity, which denotes the characterization of $\epsilon$-optimal based on primal-dual optimality gap as follows,
\begin{eqnarray}
\hbox{Gap}({\mathbf{\Sigma}}_1,{\mathbf{\Sigma}}_2 ) \!\!\!\!&:=&\!\!\!\!  {\cal{L}}_{\rho_1,\rho_2} ({\mathbf{\Sigma}}_1, {\mathbf{\Sigma}}_2^*) - {\cal{L}}_{\rho_1,\rho_2} ({\mathbf{\Sigma}}_1^*, {\mathbf{\Sigma}}_2)\nonumber\\
&\le&\!\!\!\! \epsilon.
\end{eqnarray}
Thus, it means that after $k$ iterations, we can obtain an ${\cal{O}}(1/k)$-optimal solution. The inequality (\ref{Nonergodic_IC}) calculates the optimality condition between adjacent iterations, although this can not indicate convergence, but it really can accelerate the global convergence.

The above theorem not only shows the global convergence of {{Algorithm 1}}, but also presents two cases of iteration complexity. The global convergence means that the generated sequence converges to the optimal solution based on any initial point. Further the iteration complexity results mean that how good the iteration result is after $k$ iterations. We can also find that both iteration complexity results are $O(1/k)$, which is in the same order as many first-order algorithms.
In addition, we also discuss the computational complexity in each iteration.
The main computation in each iteration comes from updating $\mathbf{W}$, $\widehat{\mathbf{W}}$, $\widetilde{\mathbf{W}}$, $\mathbf{Z}$ and the dual variables $\Lambda_1$, $\Lambda_2$, and $\Lambda_3$. The update steps of the dual variables refer to one matrix multiplication and several matrix additions whose computational complexity is ${\cal{O}}\left(n^2 d\right)$.
For updating $\mathbf{W}$, it refers to several matrix multiplications and two eigenvalue decompositions, which costs $O(n^3+n^2d+d^2n+d^3)$. For updating $\widehat{\mathbf{W}}$ and $\widetilde{\mathbf{W}}$, both of the complexities are of order $O(nd)$. Updating $\mathbf{Z}$ needs $O(n^2d)$. Therefore, the total computational complexity in each iteration is $O(n^3+n^2d+d^2n+d^3)$. In \cite{nie2013early}, the total computational complexity in each iteration is $O(n^4 + n^3d)$. Compared to \cite{nie2013early}, our method has lower complexity when the number of samples is large.

\begin{table}
\small
\begin{center}
\caption{Summary of experimental datasets. `SP', `FT', `CT' denote the number of samples, the number of features, and the number of categories, respectively.}
\label{datasets}
\begin{tabular}{|c|c|c|c|c|}
\hline
Dataset &SP   & FT  &  CT  & Type \\
\hline
{Madelon}   & {2600}           &{500}     & {2} &  {Artificial} Data   \\
\hline
{TOX-171}     & {171}             & {5748}   &{4}   &Microarray\\
\hline
{Musk}      & {476}            & {168}     & {2}         &Microarray\\
\hline
{ORL}   &{400}           & {512}     &  {40}     &{Image}\\
    \hline
{FG-NET}   & {1002}            & {907}   &{5}     & {Image}\\
   \hline
{UCF11}   & {1600}            & {512}   & {11}    &{Video} \\
   \hline
{CLL\_SUB\_111}   & {111}            &{11340}   & {3}    &{Biological Data} \\
   \hline
{HAR}   & {10299}            &{561}   & {6}    &{Smartphone  Data} \\
   \hline
\end{tabular}
\end{center}
\end{table}

\section{EXPERIMENT}
In this section, in order to sufficiently verify the effectiveness of our method, ALFS, we try to perform it on eight publicly available datasets across different domains including artificial data, microarray data, image data, video data, biological data and mobile data. As we know, datasets from diverse domains serve as a high-quality test bed for a comprehensive evaluation. In addition, these datasets are widely used for evaluating active learning or feature learning algorithms, such as Madelon \cite{Li2016Feature}, FG-NET \cite{Wang2011Active}, UCF11 \cite{Rosa2017Active}, TOX-171 \cite{Li2016Feature}, ORL \cite{Li2016Feature}, CLL\_SUB\_111 \cite{Li2016Feature}, Musk \cite{Leng2010A}, HAR \cite{Pippa2016Feature}. Table \ref{datasets} summarizes the details of the datasets used in the experiments.
\subsection{Experimental Setting}

\noindent{\textbf{Compared methods}} \ Since ALFS is related to the second group of active learning algorithms, i.e., reconstruction based methods (see discussion in the Introduction section), we compare it with some state-of-the-art approaches in this group to demonstrate the effectiveness of ALFS, including TED \cite{yu2006active}
, RRSS \cite{nie2013early}
, ALNR \cite{hu2013active}. We also compare with a randomized algorithmic CUR decomposition \cite{mahoney2009cur} that is called R-CUR. In addition, we take random sampling (RS) as another baseline. There are two variants about our method, ALFS-I and ALFS-II. ALFS-I ignores local linear reconstruction, while ALFS-II incorporates it into the learning progress.
\begin{figure*}
\centering
\subfigure[Madelon]{\includegraphics[width=0.235\linewidth]{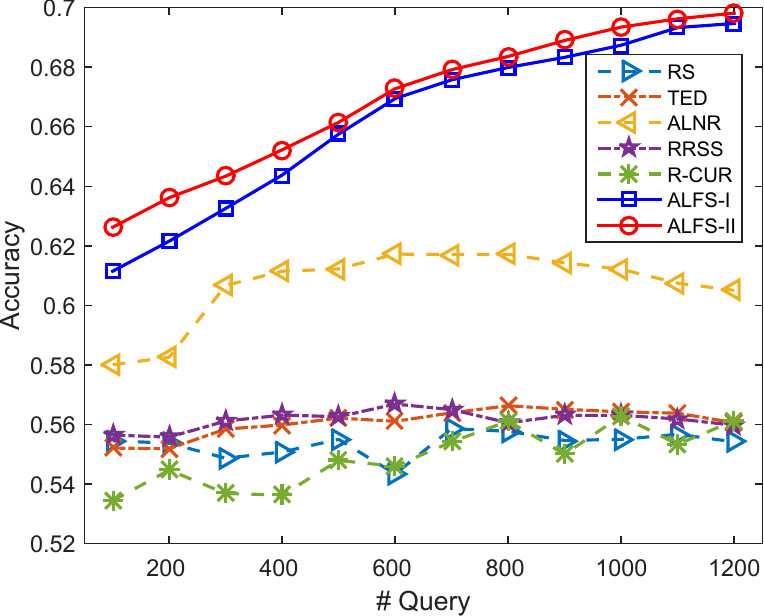}}
\subfigure[TOX-171]{\includegraphics[width=0.235\linewidth]{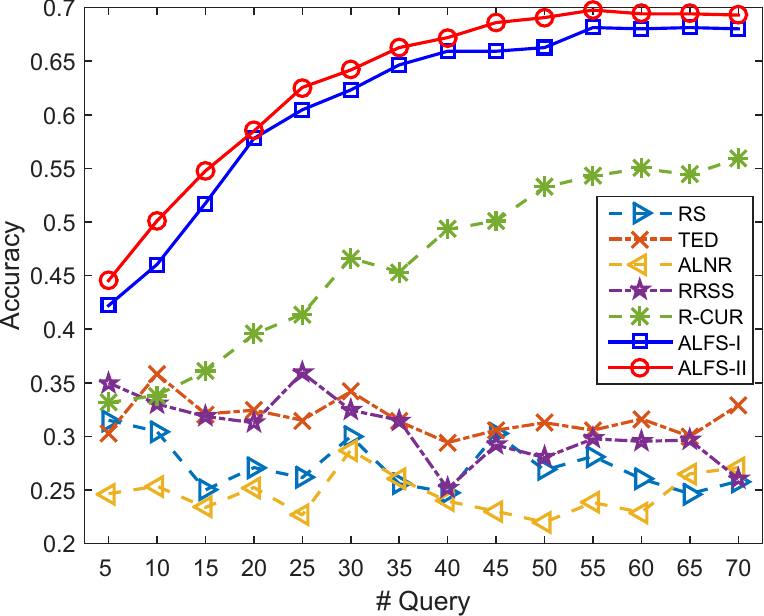}}
\subfigure[UCF11]{\includegraphics[width=0.235\linewidth]{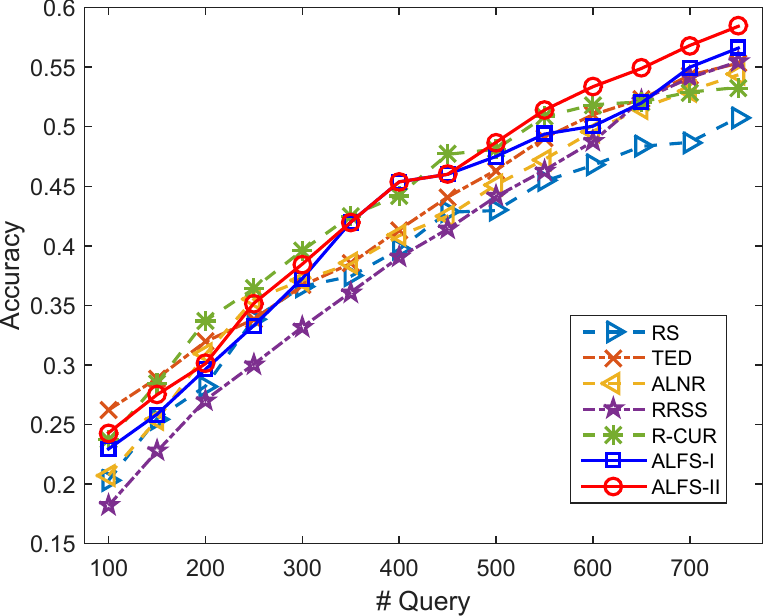}}
\subfigure[CLL\_SUB\_111]{\includegraphics[width=0.235\linewidth]{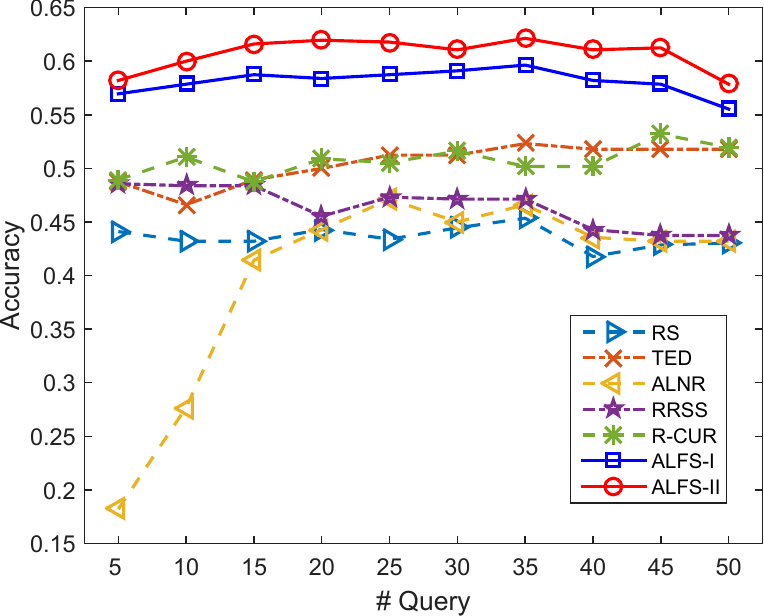}}
\subfigure[Musk]{\includegraphics[width=0.235\linewidth]{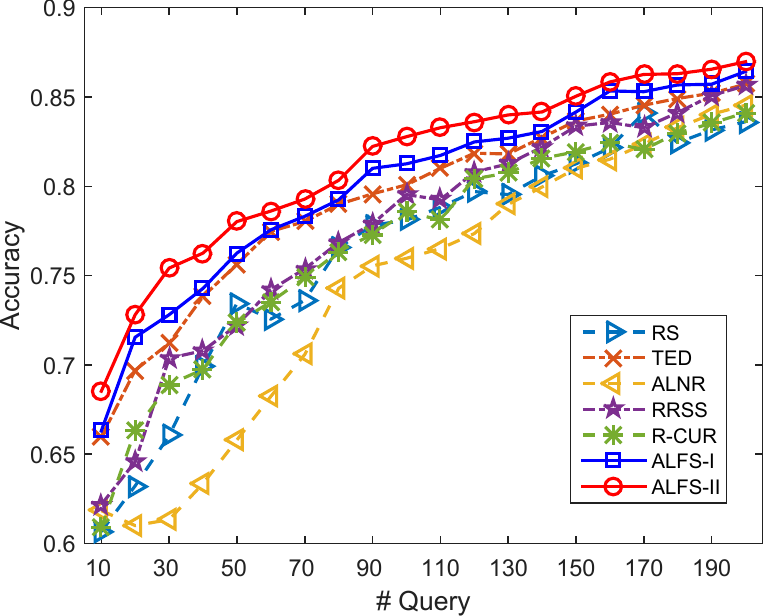}}
\subfigure[ORL]{\includegraphics[width=0.235\linewidth]{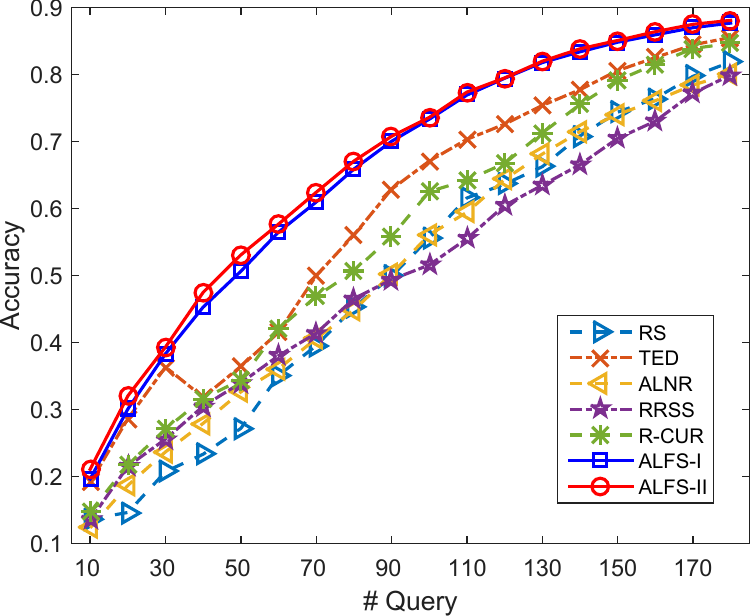}}
\subfigure[FG-NET]{\includegraphics[width=0.235\linewidth]{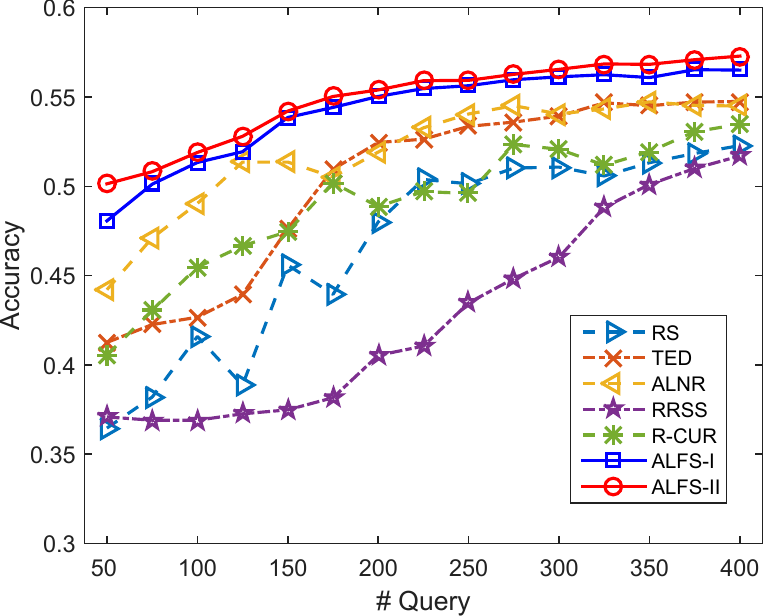}}
\subfigure[HAR]{\includegraphics[width=0.235\linewidth]{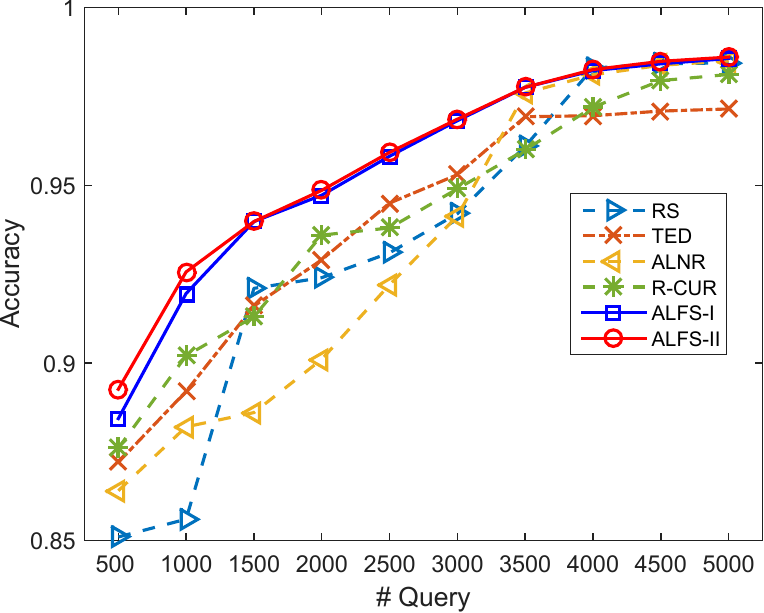}}
\vspace{-.1in}
\caption{Comparisons of different active learning methods combined with the SVM classifier on eight benchmark datasets. The curve shows the learning accuracy over queries.}
\label{query_accy1}
\end{figure*}

\begin{figure*}
\centering
\subfigure[Madelon]{\includegraphics[width=0.235\linewidth]{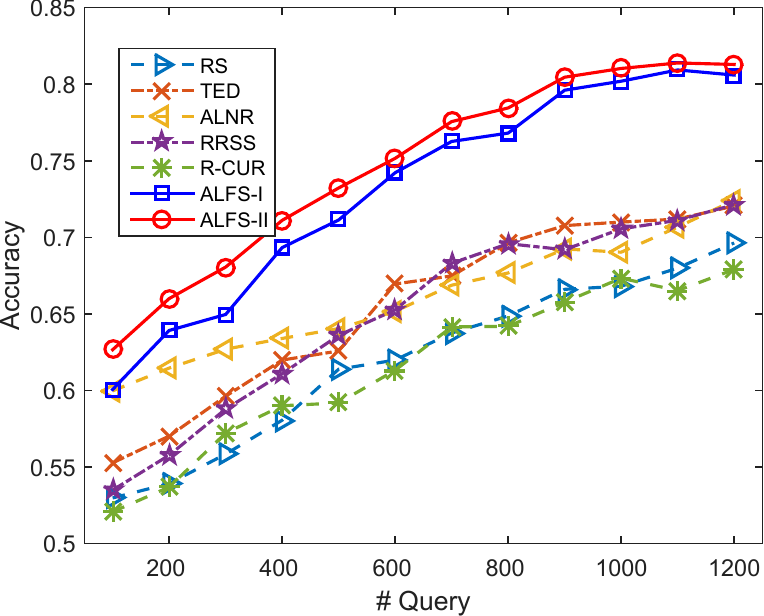}}
\subfigure[TOX-171]{\includegraphics[width=0.235\linewidth]{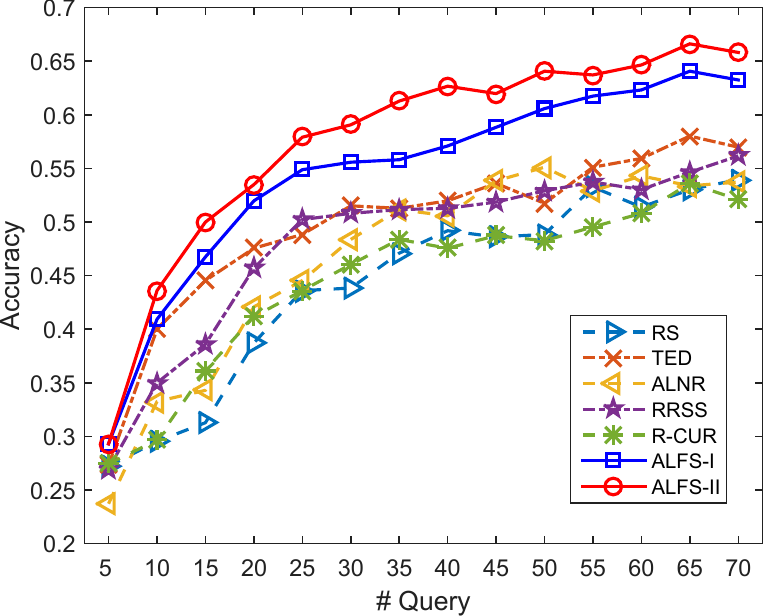}}
\subfigure[UCF11]{\includegraphics[width=0.235\linewidth]{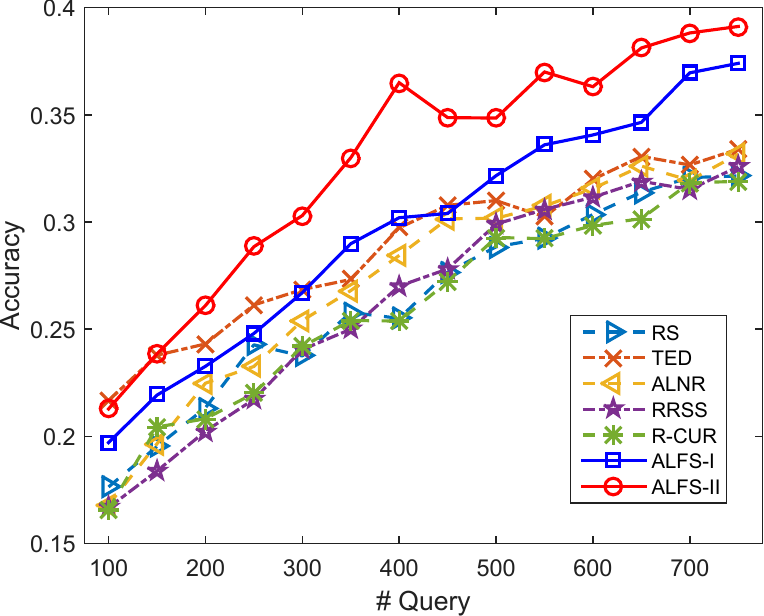}}
\subfigure[CLL\_SUB\_111]{\includegraphics[width=0.235\linewidth]{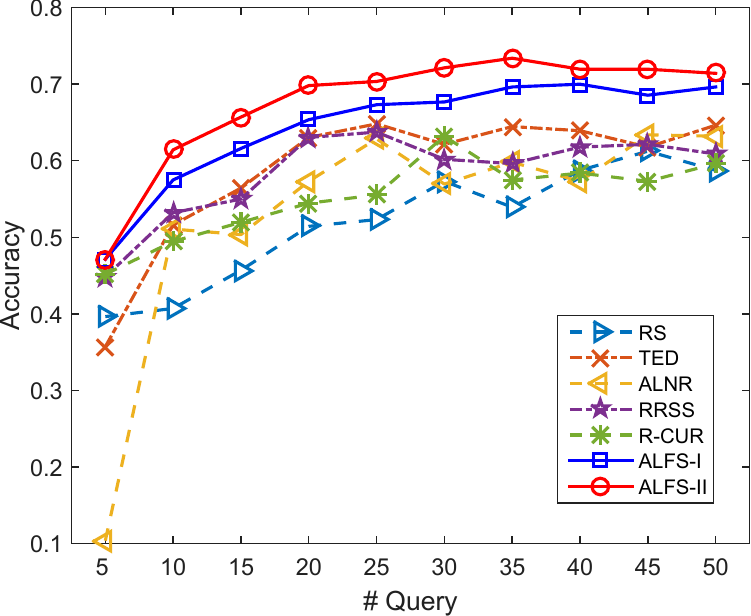}}
\subfigure[Musk]{\includegraphics[width=0.235\linewidth]{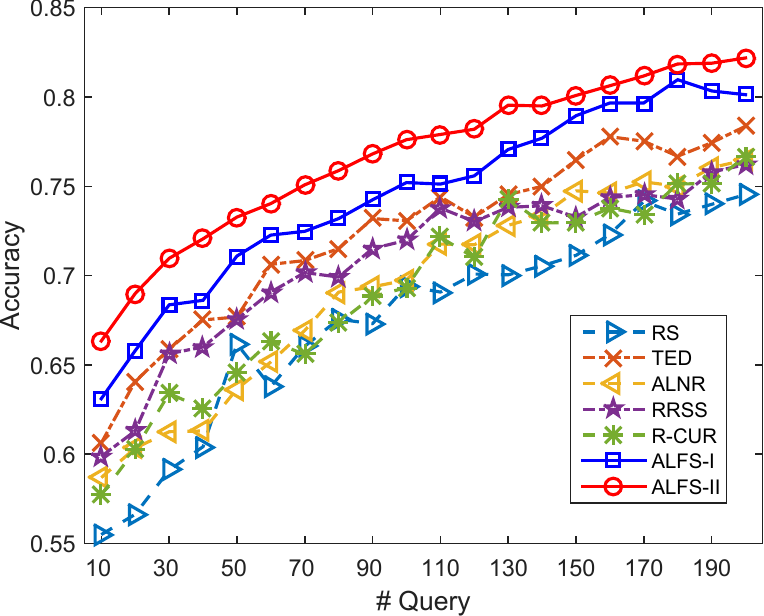}}
\subfigure[ORL]{\includegraphics[width=0.235\linewidth]{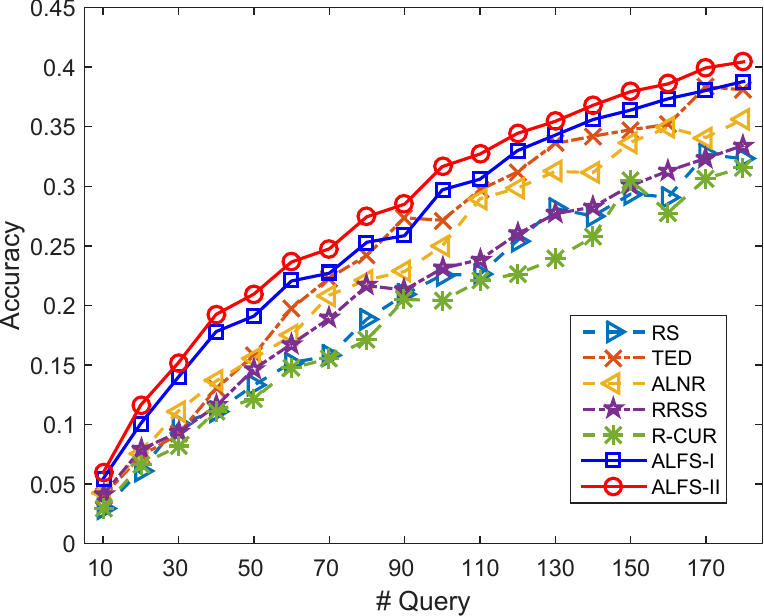}}
\subfigure[FG-NET]{\includegraphics[width=0.235\linewidth]{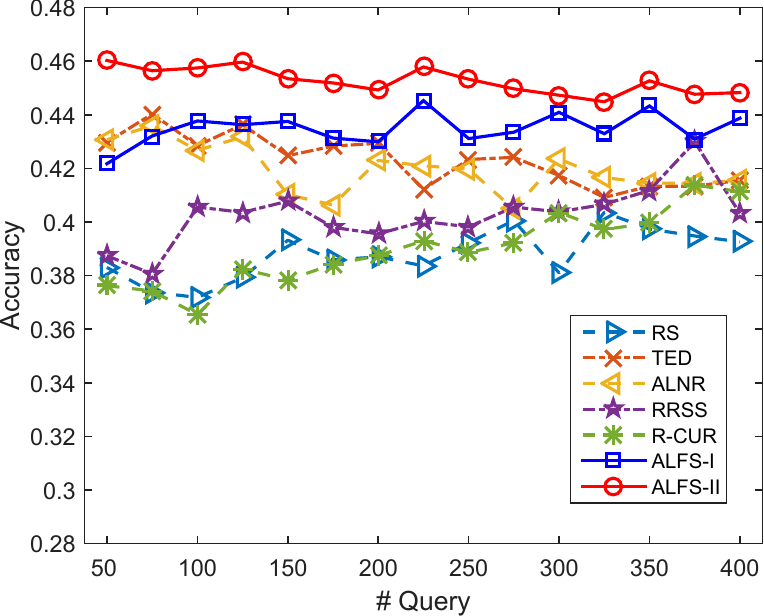}}
\subfigure[HAR]{\includegraphics[width=0.235\linewidth]{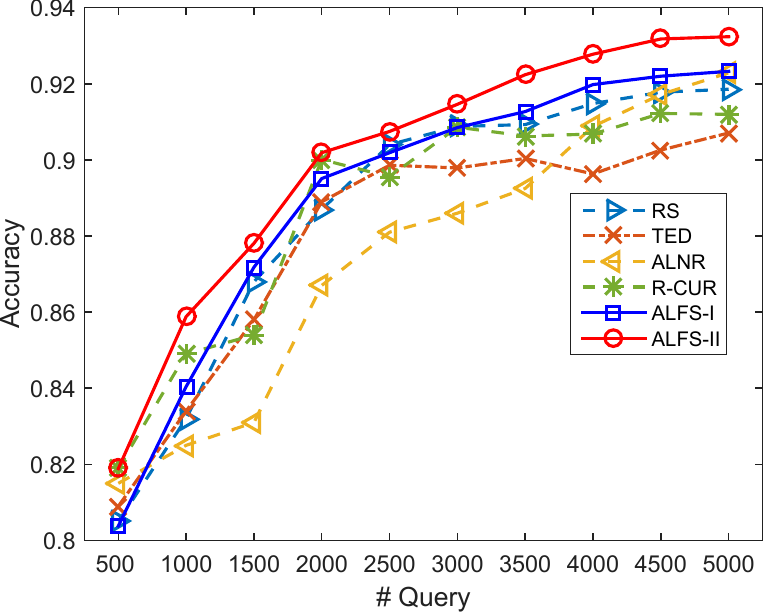}}
\vspace{-.1in}
\caption{Comparisons of different active learning methods combined with the decision tree classifier on eight benchmark datasets. The curve shows the learning accuracy over queries.}
\label{query_accy2}
\vspace{-0.15in}
\end{figure*}

To further show the benefit of simultaneous active sample selection and feature selection, we also compare our ALFS against some feature selection approaches combined with the active learning approaches above, i.e., first using feature selection methods to reduce the dimension, and then applying the active learning methods above to select samples. We use five kinds of unsupervised feature selection methods: Laplacian
\cite{he2005laplacian}, SPEC
\cite{zhao2007spectral}, UDFS \cite{Yang2011l}, Inf-FS \cite{Roffo2015Infinite}, and SOGFS
\cite{Nie2016Unsupervised}, to combine with the active learning algorithms in the experiments, respectively.

\vspace{0.03in}
\noindent \textbf{Experimental protocol} \ Following \cite{nie2013early}, for each dataset, we first randomly select 50\% of the data points as candidate samples for training, from which we apply the compared active learning methods to select a subset of samples to request human labeling.
Using the selected samples and their queried labels as training data, we learn a classification model, and evaluate the representativeness of the selected samples in terms of classification accuracy on the remaining 50\% data samples. The latter is regarded as the testing data.
In order to demonstrate that our method is not sensitive to different classifiers, we use two kinds of classical classification models: support vector machine (SVM) and decision tree, to evaluate the effectiveness of the proposed method. For simplicity, we use the linear kernel in SVM, and fix the hyperparameter $C=100$ through the experiments.
The parameters $\alpha$, $\beta$, and $\lambda$ in our algorithm  are searched from $\{10^{-4}, 10^{-3}, \ldots, 10^0, \ldots , 10^3\}$.
For a fair comparison, the parameters in TED, RRSS, and ALNR are also searched from the same space.
In the experiment, we repeat every test case 10 times, and report the average classification performance.

\begin{table*}
\scriptsize
\caption{Accuracy (\%) of feature selection + active learning algorithms on the Madelon dataset. Best results in each column are highlighted in bold fonts.}
\label{performance1}
\centering
\subtable[SVM]{
\begin{tabular}{c|c|c|c|c|c|c}
\hline
\multirow{2}{*}{Method} & \multicolumn{6}{c}{\multirow{1}{*}{\#Dim}} \\
 \cline{2-7}  & 10&  30&  50      &  70     & 90   &  500 \\
\hline
Laplacian$+$TED    &57.2    &59.4    &58.8    &58.2    &56.6    &56.1    \\
SPEC$+$TED    &54.3    &54.8    &56.0    &56.0    &54.7    &56.1    \\
SOGFS$+$TED    &49.9    &51.9    &53.1    &55.1    &57.2    &56.1    \\
UDFS$+$TED    &51.0    &51.2    &52.4    &54.7    &55.3    &56.1    \\
Inf-FS$+$TED    &56.6    &57.0    &57.0    &57.2    &57.5    &56.1    \\
\hline
Laplacian$+$RRSS    &64.3    &63.7    &61.2    &60.0    &58.7    &56.0    \\
SPEC$+$RRSS    &62.8    &59.8    &58.6    &58.0    &57.5    &56.0    \\
SOGFS$+$RRSS    &48.6    &56.2    &59.2    &60.0    &60.7    &56.0    \\
UDFS$+$RRSS    &50.0    &50.6    &54.6    &56.2    &56.4    &56.0    \\
Inf-FS$+$RRSS    &62.0    &61.2    &60.8    &60.3    &60.1    &56.0    \\
\hline
Laplacian$+$ALNR    &62.1    &61.6    &61.1    &60.6    &60.3    &\textbf{60.5}    \\
SPEC$+$ALNR    &60.2    &59.3    &58.8    &59.0    &58.9    &60.5    \\
SOGFS$+$ALNR    &49.5    &50.7    &50.4    &51.1    &59.3    &60.5    \\
UDFS$+$ALNR    &49.6    &50.6    &54.5    &56.1    &56.5    &60.5    \\
Inf-FS$+$ALNR    &61.8    &61.2    &60.7    &60.3    &60.0    &60.5    \\
\hline
R-CUR    &49.5    &54.4    &52.0    &56.1    &54.1    &55.4    \\
\hline
ALFS-I    &69.5    &64.2    &61.9    &61.5    &61.0    &56.7 \\
ALFS-II    &\textbf{69.8}    &\textbf{64.8}    &\textbf{62.7}    &\textbf{62.2}    &\textbf{61.8}    &57.2  \\
\hline
\end{tabular}
\label{tab:firsttable}
}
\qquad
\subtable[Decision Tree]{
\begin{tabular}{c|c|c|c|c|c|c}
\hline
\multirow{2}{*}{Method} & \multicolumn{6}{c}{\multirow{1}{*}{\#Dim}} \\
 \cline{2-7}  & 10&  30&  50      &  70     & 90   &  500 \\
\hline
Laplacian$+$TED    &56.4    &59.8    &63.7    &67.7    &70.6    &72.1    \\
SPEC$+$TED    &52.0    &54.8    &57.2    &61.0    &64.0    &72.1    \\
SOGFS$+$TED    &51.5    &54.1    &55.2    &58.0    &61.4    &72.1    \\
UDFS$+$TED    &51.9    &52.3    &52.8    &55.3    &56.4    &72.1    \\
Inf-FS$+$TED    &55.9    &60.4    &62.4    &65.7    &68.4    &72.1    \\
\hline
Laplacian$+$RRSS    &70.0    &78.2    &76.5    &76.1    &74.3    &72.1    \\
SPEC$+$RRSS    &66.7    &63.1    &63.3    &65.9    &65.0    &72.1    \\
SOGFS$+$RRSS    &50.9    &54.7    &60.4    &65.3    &68.3    &72.1    \\
UDFS$+$RRSS    &50.8    &52.0    &52.8    &57.0    &58.0    &72.1    \\
Inf-FS$+$RRSS    &74.5    &78.1    &76.7    &75.6    &74.7    &72.1    \\
\hline
Laplacian$+$ALNR    &70.2    &79.3    &77.0    &76.0    &75.4    &72.4    \\
SPEC$+$ALNR    &67.5    &64.9    &65.0    &66.5    &66.0    &72.4    \\
SOGFS$+$ALNR    &51.4    &51.9    &52.1    &53.1    &62.8    &72.4    \\
UDFS$+$ALNR    &51.2    &51.2    &53.6    &56.6    &57.6    &72.4    \\
Inf-FS$+$ALNR    &73.8    &77.6    &75.9    &76.0    &74.5    &72.4    \\
\hline
R-CUR    &50.3    &52.4    &52.0    &55.6    &53.7    &67.9    \\
\hline
ALFS-I    &80.6    &78.9    &77.8    &76.6    &75.3    &72.9 \\
ALFS-II    &\textbf{81.3}    &\textbf{79.9}    &\textbf{79.1}    &\textbf{78.5}    &\textbf{77.4}    &\textbf{74.7}  \\
\hline
\end{tabular}
\label{tab:secondtable}
}
\end{table*}

\begin{table*}
\scriptsize
\caption{Accuracy (\%) of feature selection + active learning algorithms on the TOX-171 dataset. Best results in each column are highlighted in bold fonts.}
\label{performance1}
\centering
\subtable[SVM]{
\begin{tabular}{c|c|c|c|c|c|c}
\hline
\multirow{2}{*}{Method} & \multicolumn{6}{c}{\multirow{1}{*}{\#Dim}} \\
 \cline{2-7}  & 10&  30&  50      &  70     & 90   &  5748 \\
\hline
Laplacian$+$TED    &48.5    &58.1    &60.8    &60.7    &60.7    &32.9    \\
SPEC$+$TED    &27.8    &30.2    &29.4    &30.0    &29.3    &32.9    \\
SOGFS$+$TED    &54.6    &60.1    &63.1    &62.4    &62.2    &32.9    \\
UDFS$+$TED    &37.0    &56.5    &61.2    &61.1    &61.6    &32.9    \\
Inf-FS$+$TED    &42.8    &47.6    &47.9    &48.7   &47.1    &32.9    \\
\hline
{Laplacian}$+${RRSS}    &52.0    &59.0    &60.6    &59.2    &59.1    &26.1    \\
{SPEC}$+${RRSS}    &25.9 &  23.3  & 23.1  & 24.9  & 25.0   &26.1    \\
SOGFS$+$RRSS    &54.6    &60.2    &62.3    &62.1    &62.5    &26.1    \\
UDFS$+$RRSS    &44.0    &57.6    &60.6    &61.5    &60.6    &26.1    \\
Inf-FS$+$RRSS    &44.7    &48.6    &48.4    &47.6    &46.7    &26.1    \\
\hline
Laplacian$+$ALNR    &51.5    &55.6    &58.4    &57.3    &56.2    &{27.1}    \\
SPEC$+$ALNR    &25.8    &26.5    &24.0    &24.9    &24.1    &27.1    \\
SOGFS$+$ALNR    &54.1    &56.6    &60.3    &57.3    &59.0    &27.1    \\
UDFS$+$ALNR    &46.9    &56.4    &60.4    &62.2   &62.1    &27.1    \\
Inf-FS$+$ALNR    &45.2    &47.8    &47.0    &45.9    &45.7    &27.1    \\
\hline
R-CUR    &40.4    &50.5    &53.4    &53.4    &55.5    &\textbf{41.3}    \\
\hline
ALFS-I    &53.7    &66.5    &67.6    &67.4    &67.6    &36.6 \\
ALFS-II    &\textbf{58.4}    &\textbf{68.0}    &\textbf{69.3}    &\textbf{69.0}    &\textbf{68.5}    &{40.7}  \\
\hline
\end{tabular}
\label{tab:firsttable}
}
\qquad
\subtable[Decision Tree]{
\begin{tabular}{c|c|c|c|c|c|c}
\hline
\multirow{2}{*}{Method} & \multicolumn{6}{c}{\multirow{1}{*}{\#Dim}} \\
 \cline{2-7}  & 10&  30&  50      &  70     & 90   &  5748 \\
\hline
Laplacian$+$TED    &45.6    &50.4    &52.7    &56.3    &57.2    &56.7    \\
SPEC$+$TED    &34.1    &39.1    &43.5   &42.4    &43.6    &56.7    \\
SOGFS$+$TED    &47.5    &50.3    &51.9    &55.4    &56.4    &56.7    \\
UDFS$+$TED    &35.2   &47.1    &50.6    &51.1    &51.6    &56.7    \\
Inf-FS$+$TED    &39.3    &51.5    &49.9    &50.7    &51.6    &56.7    \\
\hline
Laplacian$+$RRSS    &47.4    &51.5    &52.8    &54.3    &57.7    &56.2    \\
SPEC$+$RRSS    &42.2    &46.5    &48.6    &45.6   &46.3    &56.2    \\
SOGFS$+$RRSS    &45.1    &50.5    &50.5    &53.4    &53.3    &56.2    \\
UDFS$+$RRSS    &44.2    &50.1    &55.4    &55.1    &53.7    &56.2    \\
Inf-FS$+$RRSS    &45.6    &52.9   &53.5    &51.1    &55.7    &56.2    \\
\hline
Laplacian$+$ALNR    &45.6    &52.8    &50.2    &52.0    &57.9    &53.7    \\
SPEC$+$ALNR    &42.4    &45.0    &47.1    &44.9    &45.1    &53.7    \\
SOGFS$+$ALNR    &43.7    &50.3    &53.7    &52.4    &55.1    &53.7    \\
UDFS$+$ALNR    &47.0    &52.9    &54.0    &55.1    &55.1    &53.7    \\
Inf-FS$+$ALNR    &43.3    &47.0    &52.4    &51.3    &50.4    &53.7    \\
\hline
R-CUR    &43.1    &45.9    &50.8    &50.6    &51.3    &52.1    \\
\hline
ALFS-I    &56.5    &61.4    &59.7    &60.6    &63.3    &62.7 \\
ALFS-II    &\textbf{57.8}    &\textbf{62.8}    &\textbf{62.8}    &\textbf{64.3}    &\textbf{64.8}    &\textbf{65.8}  \\
\hline
\end{tabular}
\label{tab:secondtable}
}
\end{table*}
\subsection{Experimental Result}
\noindent{\textbf{Comparison with Active Learning Algorithms}}
\ In order to demonstrate the effectiveness of our ALFS in selecting representative samples, we compare ALFS with peer state-of-the-art active learning algorithms. For ALFS and R-CUR, we vary the number of selected features from 10 to 100 with an incremental step of 10 on all datasets\footnote{When the user inputs the desired number of samples $m$ and the number of features $r$, the final outputs $m'$ and $r'$ of R-CUR \cite{mahoney2009cur} may be slightly different $m$ and $r$, respectively. }, and report the best results. We do not perform RRSS on the HAR dataset, because of its extremely time cost $O(n^4+n^3d)$, where $n$ and $d$ denote the numbers of samples and features, respectively.
The results are shown in Fig. \ref{query_accy1} and Fig. \ref{query_accy2}. We can observe that ALFS-I obtains better performance than all other candidates on all tested datasets, which shows that joint active learning with feature selection is beneficial to improving classification accuracy. ALFS-II achieves the best classification performance among all datasets under different classifiers. ALFS-I and ALFS-II perform significantly better than other methods on a subset of the evaluated datasets, such as the Madelon dataset. For the Madelon dataset, when the number of the selected samples is set to 1200 and using SVM as the classifier, ALFS-I and ALFS-II obtain $14.8\%$ and $15.4\%$ relative improvement over the second best result, ALNR, respectively. When combined with decision tree, ALFS-I and ALFS-II attain $11.3\%$ and $12.3\%$ relative improvement over ALNR, respectively. In addition, ALFS-II outperforming ALFS-I means that incorporating local linear reconstruction into the process of active learning and feature selection is helpful for improving performance.
Moreover, we also observe some other interesting phenomenon. First, we note that R-CUR does not show its competency, compared to ALFS-I and ALFS-II on nearly all of the datasets. The reason is that R-CUR \cite{mahoney2009cur} is a general CUR model and adopts a randomized algorithmic approach to seek the matrices $\mathbf{C}$ and $\mathbf{R}$ for satisfying (\ref{svd1}). It does not consider it as an optimization problem, making the selected samples and features unrepresentative, which limits R-CUR to be directly applied to active learning and feature selection. Second, on the Madelon dataset and the TOX-171 dataset, the state-of-the-art active learning methods, ALNR, RRSS, TED, have poor classification performance, and do not improve the accuracy apparently as the number of the selected samples to be labeled increases, when using SVM as the final classifier. Because, for the Madelon dataset there are many noisy features (Based on the introduction in the webpage\footnote{https://archive.ics.uci.edu/ml/datasets/Madelon}, there are only 5 informative features, 15 linear combinations of these five features, and 480 distractor features having no predictive power.), and for the TOX-171 dataset, there are 5748 features, being relatively large to the number of the samples. Thus, when using these features to train SVM without dimension reduction or removing noise variables, the model is easy to overfit, which degrades the generality ability of the model. This also indicates that active learning can benefit from feature selection. In contrast, when combined with decision tree, these active learning methods obtain good results on these two datasets. The reason is that decision tree only employs a small subset of features from the decision hyperplane, which can play the role of feature selection for removing noisy or redundant features to some extent. Third, active learning methods perform better than random sampling in general, especially when combined with decision tree. This shows that it is indeed meaningful to learn to select samples for human labeling in the supervised learning scenario.

\begin{table*}
\scriptsize
\caption{Accuracy (\%) of feature selection + active learning algorithms on the UCF11 dataset. Best results in each column are highlighted in bold fonts.}
\label{performance1}
\centering
\subtable[SVM]{
\begin{tabular}{c|c|c|c|c|c|c}
\hline
\multirow{2}{*}{Method} & \multicolumn{6}{c}{\multirow{1}{*}{\#Dim}} \\
 \cline{2-7}  & 10&  30&  50      &  70     & 90   &  512 \\
\hline
Laplacian$+$TED    &29.1    &40.9    &45.2    &50.4    &55.5    &55.3    \\
SPEC$+$TED         &21.5    &33.2    &41.0    &48.6    &54.3    &55.3    \\
SOGFS$+$TED        & 29.9   &40.7    &45.1    &48.9    &55.0    &55.3    \\
UDFS$+$TED        & 20.9 & 28.2 &  32.4   &36.7  & 40.8 &55.3    \\
Inf-FS$+$TED        & 28.7 &  38.2&   42.0  & 44.5  & 45.5   &55.3    \\
\hline
Laplacian$+$RRSS   &36.7    &48.3    &51.6    &53.3    &55.5    &55.5    \\
SPEC$+$RRSS        &29.5    &40.8    &46.1    &50.3    &54.5    &55.5    \\
SOGFS$+$RRSS       &39.4    &46.7    &51.3    &53.7    &55.3    &55.5    \\
UDFS$+$RRSS       & 21.4  & 28.6  & 32.8  & 37.4   &41.4  &55.5    \\
Inf-FS$+$RRSS       &28.8  & 38.9 &  42.6  & 44.8  & 46.0  &55.5    \\
\hline
Laplacian$+$ALNR   &35.8    &47.1    &50.3    &52.5    &54.5    &54.4    \\
SPEC$+$ALNR        &29.3    &41.1    &46.5    &50.2    &53.5    &54.4    \\
SOGFS$+$ALNR       &38.1    &46.7    &49.5    &52.2    &54.1    &54.4    \\
UDFS$+$ALNR       &20.9  & 28.2  & 32.4  &36.7  &40.8  &54.4    \\
Inf-FS$+$ALNR       &28.7 &  38.2 &  42.0  & 44.5  & 45.5   &54.4    \\
\hline
R-CUR              &36.4    &47.7    &52.0    &52.5    &53.3    &53.3    \\
\hline
ALFS-I             &44.8    &51.8    &53.7    &56.3    &56.4    &56.6 \\
ALFS-II            &\textbf{45.3}    &\textbf{52.8}    &\textbf{55.3}    &\textbf{57.1}    &\textbf{58.4}    &\textbf{58.2}  \\
\hline
\end{tabular}
\label{tab:firsttable}
}
\qquad
\subtable[Decision Tree]{
\begin{tabular}{c|c|c|c|c|c|c}
\hline
\multirow{2}{*}{Method} & \multicolumn{6}{c}{\multirow{1}{*}{\#Dim}} \\
 \cline{2-7}  & 10&  30&  50      &  70     & 90   &  512 \\
\hline
Laplacian$+$TED    &25.6    &30.0    &31.5    &32.5    &33.2    &33.4    \\
SPEC$+$TED         &18.4    &25.6    &28.3    &30.5    &31.5    &33.4    \\
SOGFS$+$TED        &2.8    &29.4    &31.0    &31.8    &33.4    &33.4     \\
UDFS$+$TED        &11.8  & 17.8  & 22.2  & 24.6&   26.6  &33.4     \\
Inf-FS$+$TED        &15.6  & 22.7 &  26.5 &  29.2  & 28.9   &33.4     \\
\hline
Laplacian$+$RRSS   &35.3    &35.6    &31.6    &32.6    &32.4    &32.7     \\
SPEC$+$RRSS        &24.9    &26.8    &29.0    &30.1    &31.2    &32.7     \\
SOGFS$+$RRSS       &33.1    &33.0    &32.6    &33.1    &32.4    &32.7      \\
UDFS$+$RRSS       & 29.6  & 34.3  & 34.7   &34.1 &  34.1   &32.7      \\
Inf-FS$+$RRSS       &33.0 &  36.7 & 37.7  & 35.8  & 35.9   &32.7      \\
\hline
Laplacian$+$ALNR   &34.0    &35.0    &33.9    &32.7    &32.6    &33.2     \\
SPEC$+$ALNR        &25.1    &26.7    &28.7    &29.3    &31.8    &33.2      \\
SOGFS$+$ALNR       &33.3    &34.6    &32.8    &33.6    &32.3    &33.2      \\
UDFS$+$ALNR       &30.2  & 34.4   &33.3  & 34.9 & 34.5   &33.2      \\
Inf-FS$+$ALNR       & 32.9  & 36.2&   37.0   &35.6   &35.6   &33.2      \\
\hline
R-CUR              &29.0    &30.5    &30.9    &31.5    &31.3    &31.9      \\
\hline
ALFS-I             &37.4    &36.2    &36.1    &35.3    &35.2    &34.9      \\
ALFS-II    &\textbf{39.1}    &\textbf{39.1}    &\textbf{38.2}    &\textbf{38.0}    &\textbf{36.8}    &\textbf{36.9}  \\
\hline
\end{tabular}
\label{tab:secondtable}
}
\end{table*}

\begin{table*}
\scriptsize
\caption{Accuracy (\%) of feature selection + active learning algorithms on the CLL\_SUB\_111 dataset. Best results in each column are highlighted in bold fonts.}
\label{performance1}
\centering
\subtable[SVM]{
\begin{tabular}{c|c|c|c|c|c|c}
\hline
\multirow{2}{*}{Method} & \multicolumn{6}{c}{\multirow{1}{*}{\#Dim}} \\
 \cline{2-7}  & 10&  30&  50      &  70     & 90   &  907 \\
\hline
Laplacian$+$TED    & 51.8  & 54.5 &  56.3 &  55.2 &  54.8  & \textbf{51.8}   \\
SPEC$+$TED         &  47.3   &46.6 &  46.8  & 46.8 &  46.8   &\textbf{51.8}   \\
SOGFS$+$TED        & 44.1  & 48.0  & 49.5 &  50.2  & 51.1 &  \textbf{51.8}  \\
UDFS$+$TED        & 51.1  & 55.2  & 55.5 &  54.6  & 55.0  & \textbf{51.8}  \\
Inf-FS$+$TED        &  53.0  & 54.3 &  \textbf{56.4}   &\textbf{55.9}   &\textbf{55.5}  & \textbf{51.8}  \\
\hline
Laplacian$+$RRSS   & 49.1   & 49.5&    48.9  &  49.1    &49.5  &  43.8    \\
SPEC$+$RRSS        &42.7   & 43.0  &  43.0   & 42.9   & 43.2   & 43.8   \\
SOGFS$+$RRSS       &44.5 &   44.6 &   45.4  &  44.3  &  44.6 &   43.8   \\
UDFS$+$RRSS       &48.6   & 51.1  &  49.5  &  48.6  &  50.4 &   43.8  \\
Inf-FS$+$RRSS       &50.9  &  50.2  &  50.5  &  49.1   & 48.9 &   43.8  \\
\hline
Laplacian$+$ALNR   &47.5  & 47.3 &  47.0  & 47.7 &  47.9   &47.9 \\
SPEC$+$ALNR        & 42.7   &42.7   &42.7   &42.7  & 42.7   &47.9  \\
SOGFS$+$ALNR       &44.1   &43.4 &  43.9 &  43.9&   44.8  & 47.9   \\
UDFS$+$ALNR       &46.6   &47.3 &  45.7  & 45.9  & 46.1  & 47.9   \\
Inf-FS$+$ALNR       &49.3  & 48.0  & 47.1 &  47.0 &  47.0 &  47.9\\
\hline
R-CUR              & 50.2   & 48.4   & 49.5   & 48.8   & 47.9   & 47.0    \\
\hline
ALFS-I             &55.5  &  55.0   & 55.0  &  55.2  &  54.1   & 44.8 \\
ALFS-II            &\textbf{57.9}   & \textbf{57.0}   & 55.9  &  \textbf{55.9}  &  \textbf{55.5}  &  45.4 \\
\hline
\end{tabular}
\label{tab:firsttable}
}
\qquad
\subtable[Decision Tree]{
\begin{tabular}{c|c|c|c|c|c|c}
\hline
\multirow{2}{*}{Method} & \multicolumn{6}{c}{\multirow{1}{*}{\#Dim}} \\
 \cline{2-7}  & 10&  30&  50      &  70     & 90   &  907 \\
\hline
Laplacian$+$TED    & 51.4  & 55.4  & 55.7 &  55.9 &  55.7  & 65.9    \\
SPEC$+$TED         &48.8  & 48.8   &48.2   &48.2 &  49.3   &65.9   \\
SOGFS$+$TED        &43.2  & 47.5 &  52.0 &  54.6&   53.4  & 65.9    \\
UDFS$+$TED        &  50.9  & 56.6   &54.6  & 58.8 &  59.1  & 65.9    \\
Inf-FS$+$TED        &  53.9  & 55.9  & 56.3   &56.2  & 58.4 &  65.9   \\
\hline
Laplacian$+$RRSS   & 54.8  &  56.8  &  56.6  &  55.0  &  54.8  &  66.3    \\
SPEC$+$RRSS        &46.8   & 52.1   & 54.1   & 52.1   & 51.4  &  66.3  \\
SOGFS$+$RRSS       & 50.5  &  52.3   & 54.3  &  54.8  &  51.8  &  66.3    \\
UDFS$+$RRSS       &54.5  &  58.6   & 59.6  &  60.0   & 58.9  &  66.3   \\
Inf-FS$+$RRSS       & 52.1   & 53.2   & 54.6   & 59.6    &58.2   & 66.3    \\
\hline
Laplacian$+$ALNR   &  52.5  &  55.9 &   52.9  &  52.1 &   51.1  &  54.5   \\
SPEC$+$ALNR        &   42.5 &   47.0  &  49.8 &   49.6 &   50.2 &   54.5     \\
SOGFS$+$ALNR       &  48.6 &   48.9  &  49.6  &  49.5  &  49.5  &  54.5     \\
UDFS$+$ALNR       & 48.4  &  49.6  &  55.4 &   52.1 &   54.5 &   54.5    \\
Inf-FS$+$ALNR       &50.2  &  54.5 &   54.8  &  54.5 &   55.0   & 54.5    \\
\hline
R-CUR              & 50.5  &  57.7  &  56.3  &  56.1   & 56.1  &  62.1     \\
\hline
ALFS-I             &63.2 &  65.2  & 66.4  & 67.0  & 67.5 &  69.6      \\
ALFS-II       &\textbf{65.5} &  \textbf{68.2} &  \textbf{69.6}  & \textbf{70.5} &  \textbf{70.7} &  \textbf{71.4}  \\
\hline
\end{tabular}
\label{tab:secondtable}
}
\end{table*}

\begin{table*}
\scriptsize
\caption{Accuracy (\%) of feature selection + active learning algorithms on the Musk dataset. Best results in each column are highlighted in bold fonts.}
\label{performance1}
\centering
\subtable[SVM]{
\begin{tabular}{c|c|c|c|c|c|c}
\hline
\multirow{2}{*}{Method} & \multicolumn{6}{c}{\multirow{1}{*}{\#Dim}} \\
 \cline{2-7}  & 10&  30&  50      &  70     & 90   &  168 \\
\hline
Laplacian$+$TED    &60.3    &66.8    &71.0    &75.9    &80.3    &85.7    \\
SPEC$+$TED         &65.1    &76.2    &81.6    &83.8    &84.7    &85.7    \\
SOGFS$+$TED        &68.1    &70.6    &73.7    &78.2    &82.3    &85.7    \\
UDFS$+$TED        &58.9   &67.1   &76.2   &80.2   &81.1   &85.7 \\
Inf-FS$+$TED        &59.3    &68.4    &73.6    &76.9    &79.0  &85.7  \\
\hline
Laplacian$+$RRSS   &63.6    &67.4    &72.4    &75.6    &80.7    &85.6    \\
SPEC$+$RRSS        &71.3    &78.1    &81.1    &83.8    &84.8    &85.6    \\
SOGFS$+$RRSS       &66.9    &69.4    &75.0    &79.5    &83.2    &85.6    \\
UDFS$+$RRSS       &61.2 &  66.6  &  75.8  &  78.8 &   82.2    &85.6    \\
Inf-FS$+$RRSS       &66.4   &72.2   &75.5 &  77.7  & 79.1    &85.6    \\
\hline
Laplacian$+$ALNR   &63.1    &67.0    &73.9    &74.4    &79.5    &84.5    \\
SPEC$+$ALNR        &72.6    &78.3    &80.7    &82.3    &83.7    &84.5    \\
SOGFS$+$ALNR       &64.5    &71.6    &72.8    &75.3    &79.9    &84.5    \\
UDFS$+$ALNR       &60.5 &  72.3  & 77.2 &  81.8 &  83.1   &84.5    \\
Inf-FS$+$ALNR       &65.6 &  71.1 &  75.3  & 76.6  & 78.0    &84.5    \\
\hline
R-CUR              &71.9    &76.7    &80.8    &82.0    &83.7    &84.1    \\
\hline
ALFS-I             &74.4    &81.2    &83.9    &84.9    &85.6    &86.4 \\
ALFS-II            &\textbf{77.4}    &\textbf{83.3}    &\textbf{85.6}    &\textbf{86.2}    &\textbf{86.4}    &\textbf{87.0}  \\
\hline
\end{tabular}
\label{tab:firsttable}
}
\qquad
\subtable[Decision Tree]{
\begin{tabular}{c|c|c|c|c|c|c}
\hline
\multirow{2}{*}{Method} & \multicolumn{6}{c}{\multirow{1}{*}{\#Dim}} \\
 \cline{2-7}  & 10&  30&  50      &  70     & 90   &  168 \\
\hline
Laplacian$+$TED    &61.0    &66.6    &69.5    &75.7    &78.1    &78.4    \\
SPEC$+$TED         &62.3    &72.3    &74.8    &76.3    &77.0    &78.4    \\
SOGFS$+$TED        &71.6    &72.7    &75.1    &77.3    &78.6    &78.4     \\
UDFS$+$TED        &59.2 &  67.6 &  71.6  & 73.2  & 75.2    &78.4     \\
Inf-FS$+$TED        &  58.3  & 67.1  & 69.6 &  72.1  & 71.8   &78.4     \\
\hline
Laplacian$+$RRSS   &68.2    &71.8    &73.7    &74.7    &77.2    &76.2     \\
SPEC$+$RRSS        &71.0    &73.7    &74.5    &75.3    &77.0    &76.2     \\
SOGFS$+$RRSS       &71.5    &72.2    &74.6    &76.8    &77.0    &76.2      \\
UDFS$+$RRSS        &68.9  & 70.2 &  74.3  & 74.3 &  77.6    &76.2     \\
Inf-FS$+$RRSS        &   70.0 &  71.3 &  72.7  & 73.9  & 73.6   &76.2     \\
\hline
Laplacian$+$ALNR   &69.1    &71.5    &75.0    &74.8    &75.3    &76.5     \\
SPEC$+$ALNR        &72.1    &72.6    &75.9    &75.2    &75.0    &76.5      \\
SOGFS$+$ALNR       &72.7    &74.1    &75.8    &77.8    &76.4    &76.5      \\
UDFS$+$ALNR        &70.2  & 70.5  & 74.0 &  73.7  & 76.4   &76.5     \\
Inf-FS$+$ALNR        &67.4&  71.8 & 73.4  & 74.2  & 72.8  &76.5     \\
\hline
R-CUR              &71.1    &73.2    &72.7    &74.2    &76.7    &74.5      \\
\hline
ALFS-I             &76.0    &79.2    &79.9    &79.8    &78.6    &79.3      \\
ALFS-II    &\textbf{78.3}    &\textbf{81.6}    &\textbf{81.8}    &\textbf{81.6}    &\textbf{81.3}    &\textbf{80.8}  \\
\hline
\end{tabular}
\label{tab:secondtable}
}
\end{table*}

\begin{table*}
\scriptsize
\caption{Accuracy (\%) of feature selection + active learning algorithms on the ORL dataset. Best results in each column are highlighted in bold fonts.}
\label{performance1}
\centering
\subtable[SVM]{
\begin{tabular}{c|c|c|c|c|c|c}
\hline
\multirow{2}{*}{Method} & \multicolumn{6}{c}{\multirow{1}{*}{\#Dim}} \\
 \cline{2-7}  & 10&  30&  50      &  70     & 90   &  512 \\
\hline
Laplacian$+$TED    &22.5    &52.4    &68.5    &77.1    &81.5    &85.6    \\
SPEC$+$TED         &10.5    &16.0    &19.8    &29.2    &39.7    &85.6    \\
SOGFS$+$TED        &30.3    &57.4    &72.1    &78.2    &83.6    &85.6    \\
UDFS$+$TED        &16.4  &49.4  & 67.3 &  75.3 &  81.9   &85.6    \\
Inf-FS$+$TED        &25.4  & 62.8 &  77.9  & 83.3  & 85.8    &85.6    \\
\hline
Laplacian$+$RRSS   &36.8    &63.2    &72.5    &78.0    &82.2    &80.0    \\
SPEC$+$RRSS        &9.3    &13.4    &16.1    &26.1    &39.8    &80.0    \\
SOGFS$+$RRSS       &40.3    &68.9    &77.6    &82.1    &82.3    &80.0    \\
UDFS$+$RRSS       &40.3    &68.9    &77.6    &82.1    &82.3    &80.0    \\
Inf-FS$+$RRSS       &40.3    &68.9    &77.6    &82.1    &82.3    &80.0    \\
\hline
Laplacian$+$ALNR   &36.3    &62.3    &71.4    &74.3    &78.3    &80.2    \\
SPEC$+$ALNR        &9.8    &13.8    &17.4    &26.2    &39.2    &80.2    \\
SOGFS$+$ALNR       &48.2    &72.1    &79.1    &82.4    &83.1    &80.2    \\
UDFS$+$ALNR       & 30.8 &  64.4  &74.6&   78.8&   81.0  &80.2    \\
Inf-FS$+$ALNR       & 43.1 &  71.6  &77.9&  80.5  & 81.8  &80.2    \\
\hline
R-CUR              &27.7    &64.4    &72.4    &77.2    &82.6    &84.7    \\
\hline
ALFS-I             &55.9    &78.3    &84.2    &86.1    &87.7    &85.7 \\
ALFS-II            &\textbf{57.2}    &\textbf{79.1}    &\textbf{84.7}    &\textbf{86.2}    &\textbf{88.1}    &\textbf{86.2}  \\
\hline
\end{tabular}
\label{tab:firsttable}
}
\qquad
\subtable[Decision Tree]{
\begin{tabular}{c|c|c|c|c|c|c}
\hline
\multirow{2}{*}{Method} & \multicolumn{6}{c}{\multirow{1}{*}{\#Dim}} \\
 \cline{2-7}  & 10&  30&  50      &  70     & 90   &  512 \\
\hline
Laplacian$+$TED    &12.8    &24.9    &30.0    &33.3    &33.5    &38.1    \\
SPEC$+$TED         &7.1    &11.7    &17.3    &20.9    &25.5    &38.1    \\
SOGFS$+$TED        &15.4    &27.1    &31.5    &35.4    &36.9    &38.1     \\
UDFS$+$TED        &9.3 & 20.3  & 24.8 &  29.4 &  31.1    &38.1     \\
Inf-FS$+$TED        &11.9 & 21.8 &  26.5  & 30.7  & 32.0    &38.1     \\
\hline
Laplacian$+$RRSS   &24.7    &29.7    &32.9    &33.7    &35.3    &33.5     \\
SPEC$+$RRSS        &15.2    &18.7    &20.8    &22.5    &25.2    &33.5     \\
SOGFS$+$RRSS       &25.1    &30.8    &33.1    &34.1    &34.6    &33.5      \\
UFDS$+$RRSS       &22.3 &  27.5  & 30.8   &30.9 &  31.9   &33.5      \\
Inf-FS$+$RRSS       &23.1  &29.3 & 30.6  &33.1  & 32.3    &33.5      \\
\hline
Laplacian$+$ALNR   &24.3    &31.3    &31.5    &32.3    &34.1    &35.7     \\
SPEC$+$ALNR        &15.0    &18.3    &20.7    &22.2    &25.3    &35.7      \\
SOGFS$+$ALNR       &26.3    &32.5    &33.2    &34.6    &34.7    &35.7      \\
UDFS$+$ALNR       &23.8 &  29.4 &  31.2 &  31.5 &  33.1   &35.7      \\
Inf-FS$+$ALNR       &23.4 &  29.5  & 30.6&  32.2   & 30.7    &35.7      \\
\hline
R-CUR              &18.6    &24.4    &26.6    &29.3    &26.1    &31.6      \\
\hline
ALFS-I             &25.5    &31.7    &35.1    &34.0    &35.9    &38.8      \\
ALFS-II    &\textbf{29.8}    &\textbf{35.6}    &\textbf{37.9}    &\textbf{37.1}    &\textbf{37.2}    &\textbf{40.5}  \\
\hline
\end{tabular}
\label{tab:secondtable}
}
\end{table*}

\begin{table*}
\vspace{-0.1in}
\scriptsize
\caption{Accuracy (\%) of feature selection + active learning algorithms on the FG-NET dataset. Best results in each column are highlighted in bold fonts.}
\label{performance1}
\centering
\vspace{-0.1in}
\subtable[SVM]{
\begin{tabular}{c|c|c|c|c|c|c}
\hline
\multirow{2}{*}{Method} & \multicolumn{6}{c}{\multirow{1}{*}{\#Dim}} \\
 \cline{2-7}  & 10&  30&  50      &  70     & 90   &  907 \\
\hline
Laplacian$+$TED    &37.3    &44.8    &47.6    &53.3    &54.6    &54.8    \\
SPEC$+$TED         & 37.4   & 37.0   & 37.0   &37.2    &37.7    &54.8    \\
SOGFS$+$TED        & 32.0   &38.8    &42.6    &45.1    &47.6    &  54.8  \\
UDFS$+$TED        & 32.4  & 41.1  & 47.4 &  50.9 &  52.3   &  54.8  \\
Inf-FS$+$TED        &37.6 & 42.5&   48.6 &  52.1 &  53.6    &  54.8  \\
\hline
Laplacian$+$RRSS   &36.8    &49.8    &52.2    &54.5    &54.6    &51.7    \\
SPEC$+$RRSS        &35.8    &35.1    &35.1    &35.4    &35.0    &51.7    \\
SOGFS$+$RRSS       &41.1    &46.6    &49.4    &51.7    &52.8    & 51.7   \\
UDFS$+$RRSS       &41.5  & 46.5 & 49.4 &  51.3&  52.0   & 51.7   \\
Inf-FS$+$RRSS       &{50.5}  & 53.3 &  54.5 &  55.0  & 55.3  & 51.7   \\
\hline
Laplacian$+$ALNR   &37.3    &50.3    &52.7    &55.0    &55.1    &54.5    \\
SPEC$+$ALNR        &36.2    &36.1    &36.5    &37.0    &37.0    &54.5    \\
SOGFS$+$ALNR       &41.6    &44.2    &48.0    &50.1    &52.4    &54.5    \\
UDFS$+$ALNR       &46.9  & 50.6  & 53.2 &  54.5  &55.1  &54.5    \\
Inf-FS$+$ALNR       &\textbf{50.6}  & \textbf{53.5}  & 54.4 &  54.8&  55.6&   54.5 \\
\hline
R-CUR              &41.3    &47.1    &48.3    &50.1    &51.1    &53.5    \\
\hline
ALFS-I             &49.3    &52.3    &55.5    &56.0    &56.5    &54.7 \\
ALFS-II            &{49.6}    &{53.4}    &\textbf{55.8}    &\textbf{56.4}    &\textbf{57.3}    &\textbf{55.7}  \\
\hline
\end{tabular}
\label{tab:firsttable}
}
\qquad
\subtable[Decision Tree]{
\begin{tabular}{c|c|c|c|c|c|c}
\hline
\multirow{2}{*}{Method} & \multicolumn{6}{c}{\multirow{1}{*}{\#Dim}} \\
 \cline{2-7}  & 10&  30&  50      &  70     & 90   &  907 \\
\hline
Laplacian$+$TED    &37.7    &39.0    &38.6    &39.7    &40.1    &41.6    \\
SPEC$+$TED         &33.9    &32.4    &33.1    &35.7    &34.2    &41.6    \\
SOGFS$+$TED        &32.7    &33.1    &35.1    &37.5    &39.2    &41.6     \\
UDFS$+$TED        & 33.1  & 35.4 &  37.6  & 39.2  & 39.3    &41.6     \\
Inf-FS$+$TED        &38.1  & 37.9  & 38.0  & 37.1 &  39.6   &41.6     \\
\hline
Laplacian$+$RRSS   &30.4    &39.1    &39.6    &40.9    &41.4    &40.3     \\
SPEC$+$RRSS        & 32.1   &33.0    &33.5    &33.5    &34.5    &40.3     \\
SOGFS$+$RRSS       &34.9    &37.5    &38.5    &41.0    &39.5    &40.3      \\
UDFS$+$RRSS       &35.1 &  36.4  & 37.5 &  38.6  & 39.9    &40.3      \\
Inf-FS$+$RRSS       &40.2  & 42.3 &  41.7&   41.5   &41.7   &40.3      \\
\hline
Laplacian$+$ALNR   &32.8    &40.1    &40.3    &41.2    &42.4    &41.6     \\
SPEC$+$ALNR        & 33.1   &34.8    &34.7    & 35.1   &34.8    &41.6      \\
SOGFS$+$ALNR       & 33.2   &36.0    &37.3    &38.9    &40.2    &41.6      \\
UDFS$+$ALNR       &38.1  & 39.4 &  42.5  & 42.4  & 41.1   &41.6      \\
Inf-FS$+$ALNR       &\textbf{ 40.4}  & 42.1  & 41.9  & 41.4   &41.5  &41.6      \\
\hline
R-CUR              &32.9    &36.6    &38.0    &39.2    &38.8    &41.1      \\
\hline
ALFS-I             &39.6    &41.3    &42.1    &42.8    &43.9    &43.7      \\
ALFS-II    &{40.3}    &\textbf{43.2}    &\textbf{42.8}    &\textbf{44.1}    &\textbf{44.8}    &\textbf{44.4}  \\
\hline
\end{tabular}
\label{tab:secondtable}
}
\end{table*}

\begin{table*}
\vspace{-0.1in}
\scriptsize
\caption{Accuracy (\%) of feature selection + active learning algorithms on the HAR dataset. Best results in each column are highlighted in bold fonts.}
\label{performance1}
\centering
\vspace{-0.2in}
\subtable[SVM]{
\begin{tabular}{c|c|c|c|c|c|c}
\hline
\multirow{2}{*}{Method} & \multicolumn{6}{c}{\multirow{1}{*}{\#Dim}} \\
 \cline{2-7}  & 10&  30&  50      &  70     & 90   &  907 \\
\hline
Laplacian$+$TED    & 76.6  & 80.6 &  83.4 &  85.0  & 87.0 &  97.2  \\
SPEC$+$TED         & 69.7  & 74.4  & 78.6  & 83.1  & 83.8 &  97.2    \\
SOGFS$+$TED        & 76.4  & 82.7&   85.1 &  88.0 &  90.0  & 97.2  \\
UDFS$+$TED        & 76.1  & 79.5  & 83.9  & 87.1  & 89.2  & 97.2  \\
Inf-FS$+$TED        & 79.7 &  84.9 &  85.0  & 89.6  & 90.7 &  97.2  \\
\hline
Laplacian$+$ALNR   &78.8 &  84.2  & 87.3 &  92.4 &  92.6 &  98.5    \\
SPEC$+$ALNR        & 75.5  & 81.4  & 86.9  & 90.5  & 92.2  & 98.5  \\
SOGFS$+$ALNR       & 77.2  & 83.4  & 85.6  & 89.2 &  89.5 &  98.5    \\
UDFS$+$ALNR       & 76.5  & 82.4  & 86.1   &90.7   &91.3  & 98.5   \\
Inf-FS$+$ALNR       &80.6  & 85.1 &  86.1  & 86.6 &  87.2 &  98.5 \\
\hline
R-CUR              &71.1    &84.2    &87.5    &90.4    &95.6   &98.1    \\
\hline
ALFS-I             &78.3  & 94.1  &\textbf{95.3} &\textbf{95.7} & \textbf{96.4}  &\textbf{98.6}\\
ALFS-II            &\textbf{89.9}   &\textbf{94.6}   &\textbf{95.3} &\textbf{95.7}  &\textbf{96.4} &\textbf{98.6} \\
\hline
\end{tabular}
\label{tab:firsttable}
}
\qquad
\subtable[Decision Tree]{
\begin{tabular}{c|c|c|c|c|c|c}
\hline
\multirow{2}{*}{Method} & \multicolumn{6}{c}{\multirow{1}{*}{\#Dim}} \\
 \cline{2-7}  & 10&  30&  50      &  70     & 90   &  907 \\
\hline
Laplacian$+$TED    &71.8  & 79.3  & 84.8 &  85.2  & 86.7  & 90.4    \\
SPEC$+$TED         &67.0  & 71.4   &77.6  & 80.9  & 83.5 &  90.4   \\
SOGFS$+$TED        & 77.4 &  80.0  & 86.0 &  88.0  & 90.5  & 90.4     \\
UDFS$+$TED        & 76.6   &79.8  & 84.7 &  85.4  & 88.0 &  90.4    \\
Inf-FS$+$TED        & 80.5  & 85.5  & 88.9   &88.0 &  90.2 &  90.4   \\
\hline
Laplacian$+$ALNR   & 77.5  & 83.0&   84.2  & 88.5 &  88.6 &  92.0     \\
SPEC$+$ALNR        & 74.7  & 80.1 &  81.7 &  84.2 &  84.9  & 92.0     \\
SOGFS$+$ALNR       &80.1  & 85.4 &  85.3  & 86.5 &  88.9  & 92.0    \\
UDFS$+$ALNR       & 78.3  & 84.0  & 86.6 &  87.2 &  87.1   &92.0     \\
Inf-FS$+$ALNR       &80.8  & 83.0 &  86.8   &87.4  & 90.8  & 92.0      \\
\hline
R-CUR              &70.3    &80.3   &84.0    &84.3    &89.4    &91.2     \\
\hline
ALFS-I             &78.2  &  91.6  &  91.9  &  92.2  &  92.1   & 92.1     \\
ALFS-II       &\textbf{91.1}  &  \textbf{92.4}   & \textbf{93.1}   & \textbf{92.9} &  \textbf{93.0}   & \textbf{92.4} \\
\hline
\end{tabular}
\label{tab:secondtable}
}
\end{table*}

\noindent{\textbf{Comparison with Feature Selection + Active Learning}}
In order to demonstrate the necessity of simultaneous sample and feature selection, we compare ALFS with peer unsupervised feature selection methods combined with the active learning algorithms above. We fix the number of selected samples to the truncations as shown in Fig. \ref{query_accy1}, and test the classification accuracies with different feature dimensions. We still do not perform RRSS on the HAR dataset, due to its extremely time cost.
The results are reported in Table 2-9. We can see that when using SVM or decision tree as the classifier, both of ALFS-I and ALFS-II outperform those approaches treating sample selection and feature selection as two separate steps. Taking the Madelon dataset as an example, when the number of selected features is set to 10, ALFS-II achieves $8.0\%$ relative improvement over RRSS combined with Laplacian and SVM, $10.7\%$ relative improvement over RRSS with SPEC and SVM, $15.1\%$ relative improvement over RRSS with Laplacian and decision tree, and $20.8\%$ relative improvement over RRSS with SPEC and decision tree. This further indicates that simultaneous sample and feature selection is promising for obtaining better performance. In addition, ALFS-II achieves better results than ALFS-I under various dimensions, which comes to the same conclusion mentioned above. We also observe that our method usually has competitive results at the lower dimensions, and even has higher accuracies than using all features under most of the datasets. It also verifies that it is meaningful to simultaneously perform active learning and feature selection.

\noindent{\textbf{Coupling of Active Learning and Feature Selection}}
In order to further show the coupling of active learning and feature selection, i.e., noisy and redundant features can bring adverse effect on sample selection, while representative samples will be beneficial to feature selection, we conduct deep studies on the TOX-171 dataset. We first show the benefit to active learning through embedding feature selection. The results are listed in Table \ref{tab:firsttable} and Table \ref{tab:secondtable}. In Table \ref{tab:firsttable}, when fixing the number of the queries, the performance using a small subset of all the features is always better than that of using all the features. For Table \ref{tab:secondtable}, the accuracies of using a subset of all the features are superior to those of using all the features under most of the cases. Even though it is not higher, the performance of using a feature subset is still comparable to that of using all the features. Therefore, it is clear that embedding feature selection is good for learning representative samples.

\begin{table*}
\vspace{-0.1in}
\caption{Results (\%) showing feature selection being good for learning representative samples on the TOX-171 dataset. Best results in each column are highlighted in bold fonts.}
\scriptsize
\label{performance1}
\centering
\vspace{-0.2in}
\subtable[SVM]{
\begin{tabular}{c|c|c|c|c|c|c|c}
\hline
\multirow{2}{*}{\#Dim} & \multicolumn{7}{c}{\multirow{1}{*}{\#Query}} \\
 \cline{2-8} & 10        &  20    &  30    & 40  & 50&60 &70 \\
\hline
10    &44.4    &51.2    &52.8    &54.0    &56.4    &57.7    &58.4\\\hline
20    &46.5    &53.0    &57.4    &61.9    &62.2    &63.4    &63.5\\\hline
30    &47.3    &56.1    &58.7    &63.7    &65.1    &67.2    &68.0\\\hline
40    &48.6    &56.4    &59.8    &64.8    &67.1    &67.3    &68.1\\\hline
50    &49.0    &57.6    &61.7    &65.2    &67.4    &68.5    &\textbf{69.3}\\\hline
60    &\textbf{50.1}    &57.7    &64.1    &66.2    &68.4    &68.8    &68.4\\\hline
70    &48.7    &58.4    &63.4    &66.6    &68.0    &68.8    &69.0\\\hline
80    &48.0    &57.8    &63.3    &66.4    &68.0    &\textbf{69.4}    &68.1\\\hline
90    &47.9    &57.9    &63.8    &\textbf{67.2}    &\textbf{69.1}    &69.2    &69.0\\\hline
100    &47.8    &\textbf{58.5}    &\textbf{64.2}    &66.4    &68.6    &69.0    &68.6\\\hline
5748    &40.8    &46.7    &45.2    &46.9    &43.5    &40.6    &40.7\\
\hline
\end{tabular}
\label{tab:firsttable}
}
\qquad
\subtable[Decision Tree]{
\begin{tabular}{c|c|c|c|c|c|c|c}
\hline
\multirow{2}{*}{\#Dim} & \multicolumn{7}{c}{\multirow{1}{*}{\#Query}} \\
 \cline{2-8} & 10        &  20    &  30    & 40  & 50&60 &70 \\
\hline
10    &41.5    &48.0    &50.8    &55.0    &56.5    &57.6    &57.8\\\hline
20    &42.6    &50.4    &54.5    &58.0    &59.8    &60.7    &63.0\\\hline
30    &\textbf{43.6}    &51.1    &56.1    &58.8    &61.3    &62.4    &62.8\\\hline
40    &\textbf{43.6}    &52.7    &55.8    &59.5    &61.3    &61.7    &63.0\\\hline
50    &43.0    &51.3    &55.9    &60.7    &62.0    &62.1    &62.8\\\hline
60    &42.3    &52.1    &55.5    &60.9    &61.7    &63.1    &63.6\\\hline
70    &42.8    &52.7    &56.7    &62.2    &62.4    &63.4    &64.3\\\hline
80    &42.8    &52.8    &56.9    &\textbf{62.7}    &63.4    &64.3    &64.5\\\hline
90    &42.8    &52.8    &56.4    &61.7    &63.1    &64.0    &64.9\\\hline
100    &42.8    &52.8    &57.6    &61.7    &\textbf{64.1}    &\textbf{64.7}    &64.8\\\hline
5748    &43.3    &\textbf{53.5}    &\textbf{59.1}    &61.4    &63.4    &64.5    &\textbf{65.8}\\
\hline
\end{tabular}
\label{tab:secondtable}
}
\end{table*}

\begin{table*}
\vspace{-0.1in}
\caption{Results (\%) showing active leaning being helpful for selecting informative features on the extended TOX-171 dataset.  Best results in each column are highlighted in bold fonts.}
\scriptsize
\label{performance2}
\centering
\vspace{-0.2in}
\subtable[SVM]{
\begin{tabular}{c|c|c|c|c|c|c}
\hline
\multirow{2}{*}{\#Query} & \multicolumn{6}{c}{\multirow{1}{*}{\#Dim}} \\
 \cline{2-7} & 10        &  30    &  50    & 70  & 90& 5748 \\
\hline
10    &41.8    &41.8    &39.8    &41.8    &43.7    &41.8\\\hline
20    &44.7    &49.5    &49.5    &48.5    &48.5    &37.9\\\hline
30    &\textbf{54.4}    &51.5    &54.4    &54.4    &53.4    &40.8\\\hline
40    &49.5    &54.4    &53.4    &54.4    &55.3    &40.8\\\hline
50    &{50.5}    &\textbf{57.3}    &59.2    &59.2    &57.3    &\textbf{46.6}\\\hline
60    &49.5    &55.3    &58.3    &58.3    &\textbf{58.3}    &39.8\\\hline
70    &48.5    &\textbf{57.3}    &\textbf{60.2}    &\textbf{60.2}    &\textbf{58.3}    &42.7\\\hline
80    &50.5    &\textbf{57.3}    &58.3    &55.3    &\textbf{58.3}    &41.8\\\hline
90    &46.6    &54.4    &58.3    &57.3    &54.4    &27.2\\\hline
100    &46.6    &53.4    &57.3    &56.3    &53.4    &27.2\\\hline
102    &47.6    &53.4    &58.3    &57.3    &52.4    &24.3\\
\hline
\end{tabular}
\label{tab:threetable}
}
\qquad
\subtable[Decision Tree]{
\begin{tabular}{c|c|c|c|c|c|c}
\hline
\multirow{2}{*}{\#Query} & \multicolumn{6}{c}{\multirow{1}{*}{\#Dim}} \\
 \cline{2-7} & 10        &  30    &  50    & 70  & 90& 5748 \\
\hline
10    &36.9    &35.9    &37.9    &37.9    &37.9    &37.9\\\hline
20    &43.7    &43.7    &43.7    &44.7    &45.6    &42.7\\\hline
30    &44.7    &49.5    &48.5    &48.5    &50.5    &55.3\\\hline
40    &47.6    &50.5    &54.4    &55.3    &55.3    &54.4\\\hline
50    &53.4    &50.5    &55.3    &\textbf{58.3}    &58.3    &56.3\\\hline
60    &51.5    &52.4    &52.4    &55.3    &53.4    &58.3\\\hline
70    &\textbf{57.3}    &\textbf{54.4}    &\textbf{57.3}    &55.3    &58.3    &\textbf{59.2}\\\hline
80    &47.6    &48.5    &50.5    &51.5    &52.4    &\textbf{59.2}\\\hline
90    &49.5    &53.4    &50.5    &50.5    &53.4    &\textbf{59.2}\\\hline
100    &51.5    &49.5    &52.4    &56.3    &\textbf{60.2}    &55.3\\\hline
102    &51.5    &49.5    &51.5    &53.4    &56.3    &56.3\\
\hline
\end{tabular}
\label{tab:fourtable}
}
\end{table*}

Next, we will demonstrate that active learning is also helpful to learning informative features. In order to make our experiments more practical and challenging, we add 20\% noisy data samples into the original dataset to form a new dataset. The noisy variable is sampled from the standard normal distribution, and the noisy label is drawn from the discrete uniform distribution on $[1,2,3,4]$. Based on the new dataset, we randomly divide it into two parts: one part is used as the candidate set to query representative samples for training, and the other part is used as the testing set. When querying the selected samples, we use SVM and decision tree as the final classifier for classifying the testing data, respectively. Table \ref{tab:threetable} and Table \ref{tab:fourtable} report the results. With the fixed feature dimensions, querying all the samples, i.e., 102 samples, for training can not obtain the best classification performance. In contrast, only selecting a small subset of samples for requesting human labeling can significantly improve classification accuracy, compared with querying all the samples. This indicates that representative samples are beneficial to learning informative features.

\subsection{CPU Time and Sensitivity Analysis}
\begin{figure}
\centering
\includegraphics[width=0.5\linewidth]{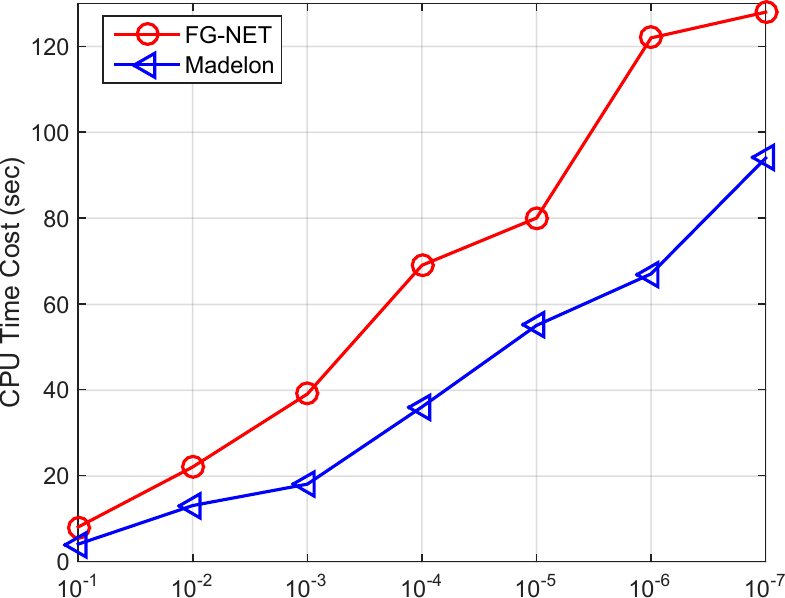}
\vspace{-.1in}
\caption{CPU time vs. convergence tolerance $\epsilon$.}
\label{time}
\vspace{-.1in}
\end{figure}
We test the CPU running time with different convergence tolerance $\epsilon$ on the Madelon dataset and the FG-NET dataset. The experiments are conducted on a laptop with Intel(R)-Core(TM) CPUs of 3.20 GHz and 4 GB RAM, and ALFS-II is implemented using MATLAB R2014b 64-bit edition without parallel operation.
The result is shown in Fig. \ref{time}. The CPU time grows linearly with $\epsilon$ increasing on both datasets.

We also study the sensitivity of our algorithm to the parameters, $\alpha$, $\beta$, and $\lambda$, on the Madelon and FG-NET datasets. In the experiment, we first fix the number of the selected features to 10, and set the number of the selected samples to the truncations as shown in Fig. \ref{query_accy1}.  Then, we fix one parameter and vary the other two parameters. We report the accuracy of our algorithm with SVM as the final classifier. The results are shown in Fig. \ref{parameter}. We can see that our method is not sensitive to all the parameters with wide ranges. In Fig. \ref{parameter}(a) and (d), when $\alpha$ and $\beta$ are set to small values, the performance of the model degrades significantly. This is because the smaller $\alpha$ and $\beta$ are, the lower the weight of the second and the third terms in (\ref{obj5}) is. In this way, it is hard to guarantee that the columns and the rows of the matrix $\mathbf{W}$ are sparse, which makes our algorithm fail to learn representative samples and features. Therefore, we should set larger $\alpha$ and $\beta$ in practice.

\begin{figure}[htb]
\centering
\subfigure[]{\includegraphics[width=0.28\linewidth]{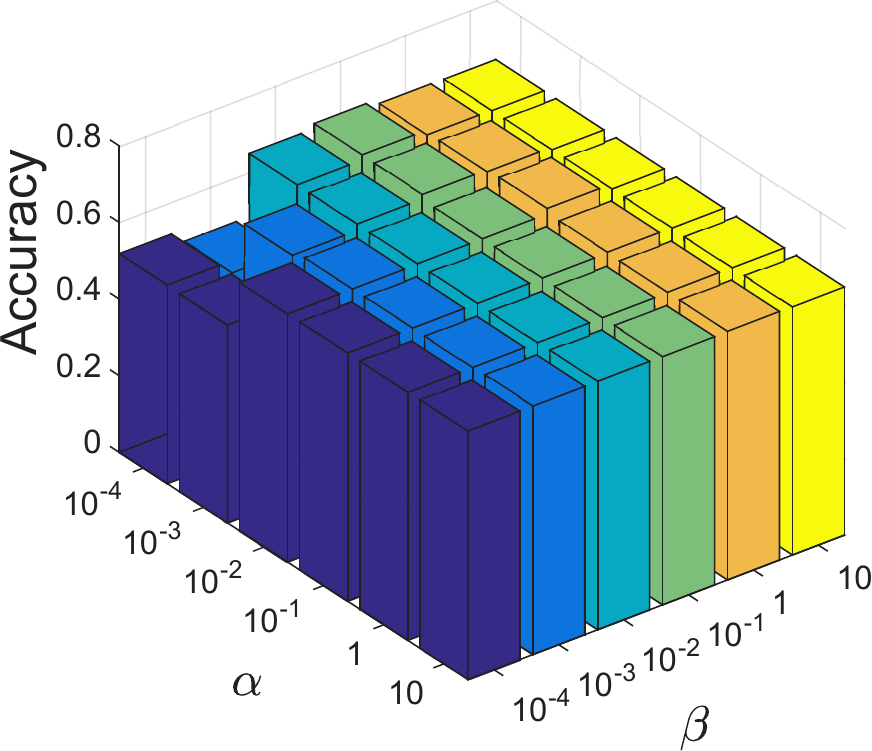}}
\subfigure[]{\includegraphics[width=0.28\linewidth]{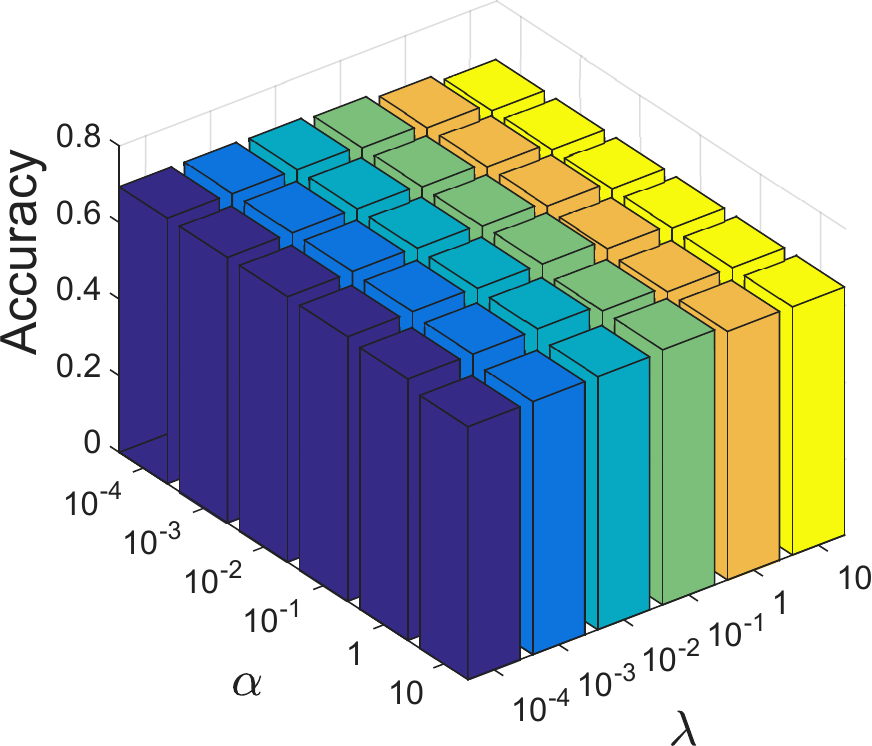}}
\subfigure[]{\includegraphics[width=0.28\linewidth]{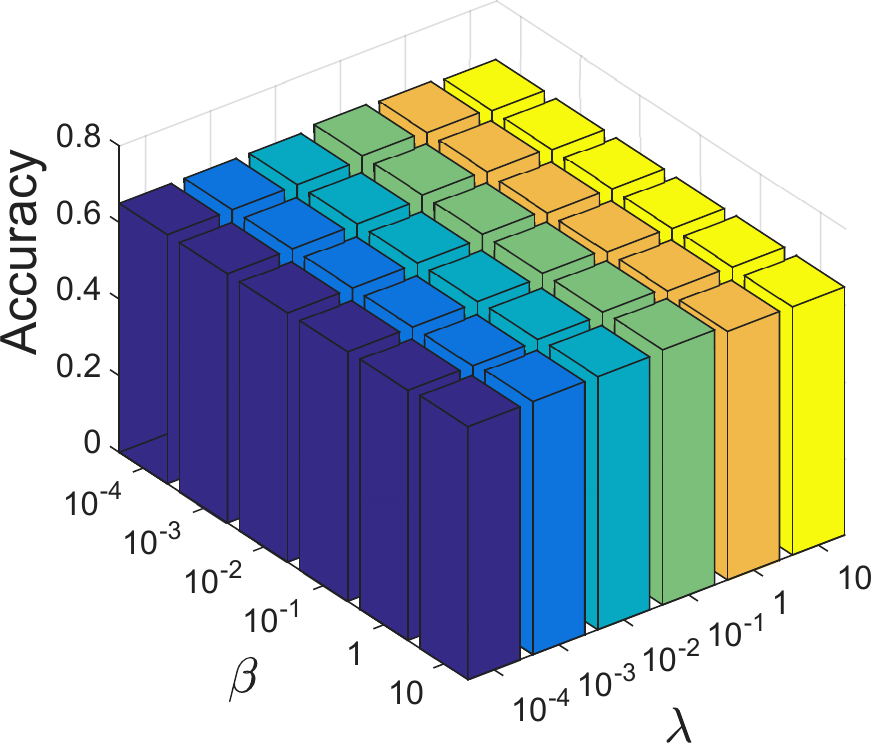}}
\subfigure[]{\includegraphics[width=0.3\linewidth]{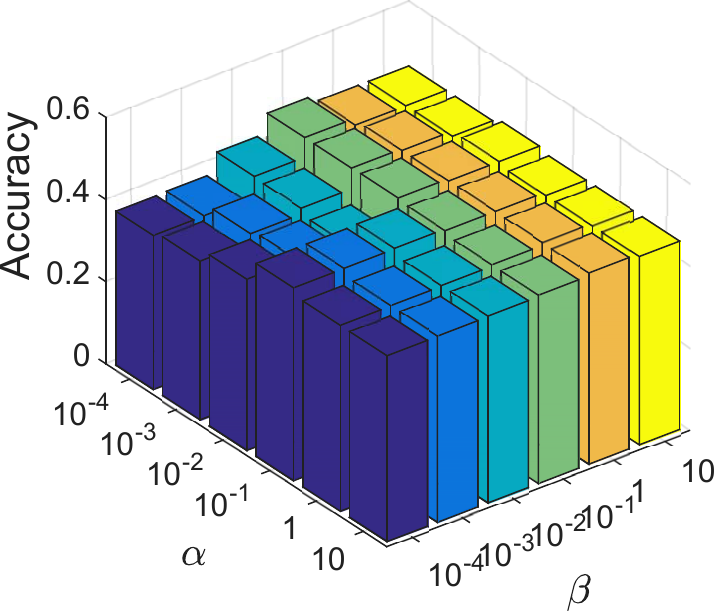}}
\subfigure[]{\includegraphics[width=0.3\linewidth]{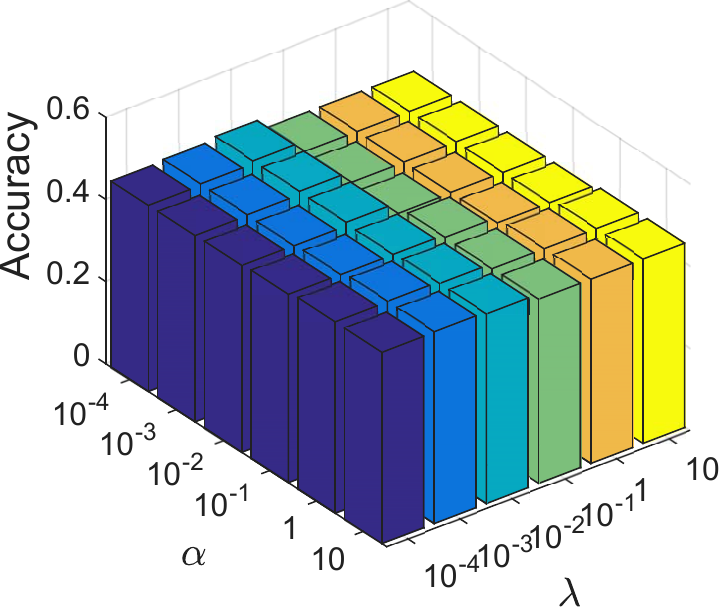}}
\subfigure[]{\includegraphics[width=0.3\linewidth]{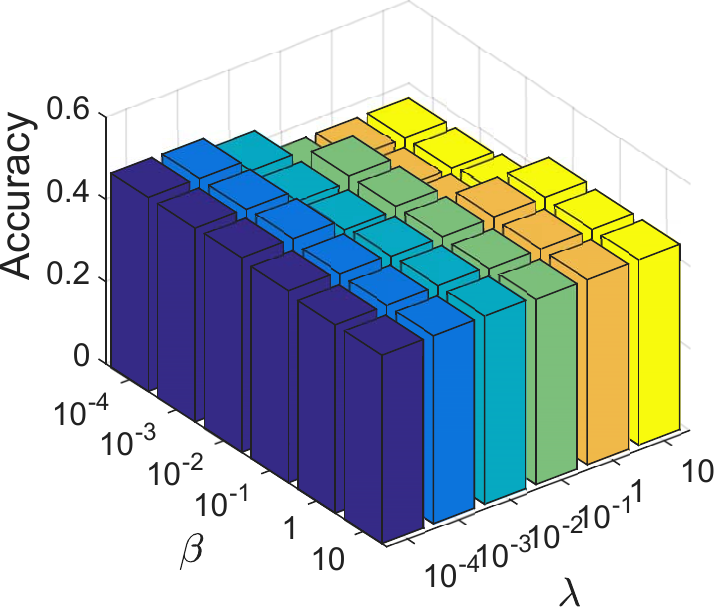}}
\vspace{-.1in}
\caption{Sensitivity study of the parameters on the Madelon and FG-NET datasets, respectively. (a),(b),(c) for the Madelon dataset, (d),(e),(f) for the FG-NET dataset. }
\label{parameter}
\vspace{-.1in}
\end{figure}

\section{CONCLUSIONS AND FUTURE WORK}
In this paper, we present a unified framework to simultaneously conduct active sample learning and feature selection (ALFS). Given an unlabeled dataset, our formulation naturally and effectively incorporates feature and sample selection by solving a regularized optimization problem rooted from CUR factorization. We further relax the original NP-hard non-convex problem into a convex one by introducing the structured sparsity-inducing norms, which allows for an efficient iterative optimization algorithm (ADMM). The superior performance of our method over the state-of-the-art methods is verified by extensive experimental evaluations with eight benchmark datasets.

Several interesting directions can be followed up, which are not covered by our current work:
\begin{itemize}
\item
{\bf{Leveraging labeled samples}}: ALFS selects samples and features from a perspective of data reconstruction in an unsupervised setting. If label information is available, we can incorporate such prior information into our framework, e.g., taking the objective function of \cite{nie2008trace} as a regularization term. This would be helpful if a specific task is only relevant to a few features and our `blind' feature selection method may keep unnecessary features although they are indispensable to represent the sample set itself.
\item
{\bf{Additional regularization terms}}:  In our work, motivated by the local reconstruction philosophy, we add the cross-sample regularization term as presented in Sect. 3.3. This term alleviates the under-determination condition of the factorization problem, and contributes to the robustness of our method. Symmetrically, a cross-feature regularization term can be also applied.
\end{itemize}

\section{acknowledgment}
The authors are thankful to Lance Warren Feagan with IBM Research - China, who helps editing and revising to improve the writing of this paper. Part of the work was done when the first author worked in IBM Research - China.
This work was partially supported by National Natural Science Foundation of China Grant No. 61532009, 11501210, 61602176. It also was partially supported by the Key Program of Shanghai Science and Technology Commission Grant No. 15JC1401700.

\ifCLASSOPTIONcaptionsoff
  \newpage
\fi

\bibliographystyle{IEEEtran}
\bibliography{sigproc}  

\begin{IEEEbiography}[{\includegraphics[width=1in,height=1.25in,clip,keepaspectratio]{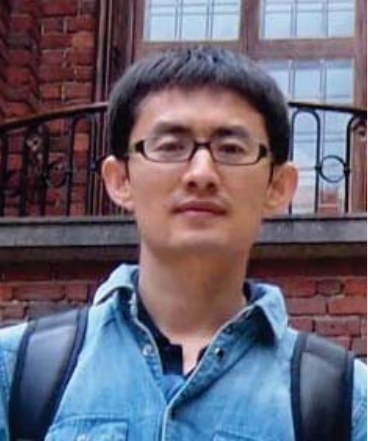}}]{Changsheng Li}
is currently a Full Research Professor from  the University of Electronic Science and Technology of China (UESTC) since 2017. He received his Ph.D. degree in pattern recognition and intelligent system from the Institute of Automation, Chinese Academy of Sciences in 2013.   After obtaining his PhD, he once worked with IBM Research-China and iDST, Alibaba Group, respectively. His research interests include machine learning, data mining.
Dr. Li has more than 30 refereed publications in international journals and conferences, including T-PAMI, T-NNLS, T-IP, T-C, PR, CVPR, AAAI, IJCAI, CIKM, MM, ICMR, etc.
\end{IEEEbiography}

\begin{IEEEbiography}[{\includegraphics[width=1in,height=1.25in,clip,keepaspectratio]{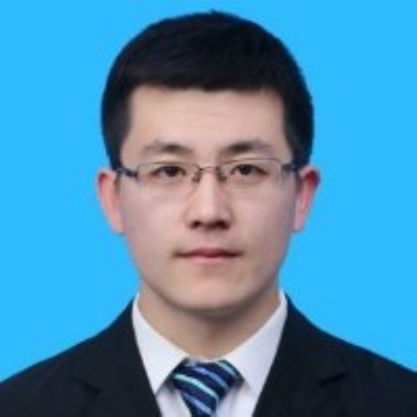}}]{Xiangfeng Wang} received the B.S. degree in mathematics and applied mathematics and the Ph.D. degree in computational mathematics, both from Nanjing University, Nanjing, China, in 2009 and 2014, respectively.
He is currently an Assistant Professor with School of Computer Science and Software Engineering, East China Normal University, Shanghai, China.
His research interests include large-scale optimization and applications on machine learning.
\end{IEEEbiography}\vspace{-0.5in}

\begin{IEEEbiography}[{\includegraphics[width=1in,height=1.25in,clip,keepaspectratio]{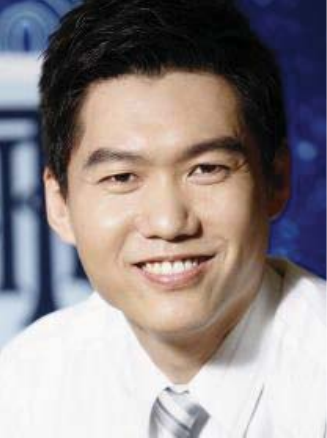}}]{Weishan Dong}
is currently a technical leader in Baidu Search.
He received his B.E. degree in computer science and technology from the University of Science and Technology of China (USTC) in 2004, and his Ph.D. degree in pattern recognition and intelligent system from the Institute of Automation, Chinese Academy of Sciences in 2009. His current research mainly focuses on data mining, especially mining big spatiotemporal data (e.g., location data and mobility data) with addressing large scale and low latency.  Dr. Dong has more than 30 refereed publications in international journals and conferences and over 20 inventions/patent applications.
\end{IEEEbiography}\vspace{-0.4in}

\begin{IEEEbiography}[{\includegraphics[width=1in,height=1.25in,clip,keepaspectratio]{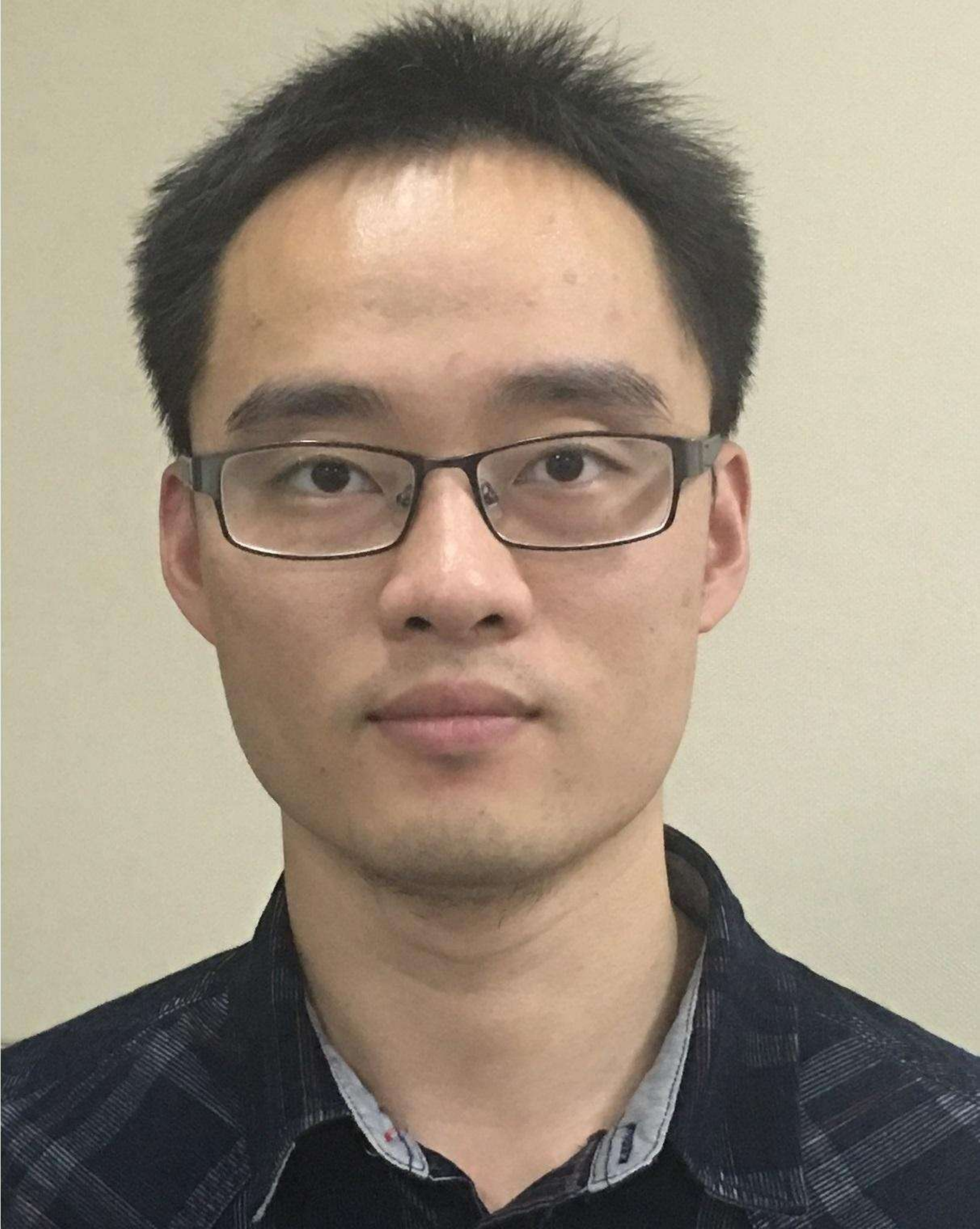}}]{Junchi Yan}
is an Associate Professor from Shanghai Jiao Tong University (SJTU). Before Joining SJTU, he worked with IBM Research-China. He has been entitled as IBM Master Inventor, and the recipient of IBM Research Division Award, China Computer Federation Doctoral Dissertation Award, and the ACM China Doctoral Dissertation Nomination Award. His research interests are computer vision and machine learning. His first author papers have appeared in TPAMI/TIP/TCYB/CVPR/ICCV/ECCV/AAAI/IJCAI. He is the Associate Editor for IEEE ACCESS and an IBM Master Inventor.
\end{IEEEbiography}\vspace{-0.4in}

\begin{IEEEbiography}[{\includegraphics[width=1in,height=1.25in,clip,keepaspectratio]{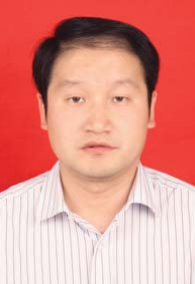}}]{Qingshan Liu}
(M'05-SM'07) is a Professor with the School of Information and Control Engineering, Nanjing University of Information Science $\&$
Technology, Nanjing, China. He received the Ph.D. degree from the National Laboratory of Pattern Recognition, Chinese Academic of Science, Beijing, China, in 2003, and the M.S. degree from the Department of Auto Control, Southeast University, Nanjing, in 2000. He was an Assistant Research Professor with the Department of Computer Science, University of Rutgers from 2010 to 2011. Before he joined Rutgers University, he was an Associate Professor with the National Laboratory of Pattern Recognition, Chinese Academic of Science. He was a recipient of the President Scholarship of the Chinese Academy of Sciences in 2003. His research interests are image and vision analysis, machine learning.
\end{IEEEbiography}\vspace{-0.4in}

\begin{IEEEbiography}[{\includegraphics[width=1in,height=1.25in,clip,keepaspectratio]{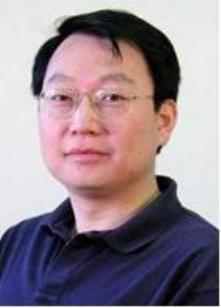}}]{Hongyuan Zha}
received the PhD degree in scientific computing from Stanford University in 1993. He is a professor at East China Normal University and the School of Computational Science and Engineering, College of Computing, Georgia Institute of Technology. Since then he has been working on information retrieval, machine learning applications, and numerical methods. He received the Leslie Fox Prize (1991, second prize) of the Institute of Mathematics and
its Applications, the Outstanding Paper Awards of the 26th International Conference on Advances in Neural Information Processing Systems (NIPS 2013), and the Best Student Paper Award (advisor) of the 34th ACM SIGIR International Conference on Information Retrieval (SIGIR 2011). He serves as an associate editor of IEEE Transactions on Knowledge and Data Engineering.
\end{IEEEbiography}

\end{document}